\documentclass{article}
\usepackage[margin=1in]{geometry}
\usepackage{booktabs} 
\usepackage{graphicx}
\usepackage{subfigure}
\usepackage{amsthm}
\usepackage{amsmath}
\usepackage{multirow}
\usepackage{enumerate}
\usepackage{amsfonts}
\usepackage{color, comment}
\usepackage{authblk}
\usepackage{algorithm, algpseudocode}

\begin{document}
\title{REMIX: Automated Exploration for Interactive Outlier Detection}

\author[1]{Yanjie Fu\thanks{fuyan@mst.edu}}
\author[2]{Charu Aggarwal\thanks{charu@us.ibm.com}}
\author[2]{Srinivasan Parthasarathy\thanks{spartha@us.ibm.com}}
\author[2]{Deepak S. Turaga\thanks{turaga@us.ibm.com}}
\author[3]{Hui Xiong\thanks{hxiong@rutgers.edu}}
\affil[1]{Missouri U. of Science \& Technology, Missouri, USA}
\affil[2]{IBM T. J. Watson Research Center, NY, USA}
\affil[3]{Rutgers University, NJ, USA}

\renewcommand\Authands{ and }
\renewcommand{\algorithmicrequire}{\textbf{Input:}}
\renewcommand{\algorithmicensure}{\textbf{Output:}}
\newtheorem{theorem}{Theorem}
\newtheorem{observation}[theorem]{Observation}

\maketitle

\begin{abstract}
Outlier detection is the identification of points in a dataset
that do not conform to the norm. Outlier detection is highly sensitive to the choice of
the detection algorithm and the feature subspace used by the algorithm. Extracting
domain-relevant insights from outliers needs systematic exploration of these choices since diverse outlier sets
could lead to complementary insights. This challenge is especially acute in an \textit{interactive setting},
where the choices must be explored in a time-constrained manner.

In this work, we present REMIX, \textit{the first system to address the problem of outlier detection in an interactive setting}.
REMIX uses a novel mixed integer programming (MIP) formulation for
\textit{automatically selecting and executing a diverse set of outlier detectors within a time limit}.
This formulation incorporates multiple aspects such as
(i) an \textit{upper limit on the total execution time of detectors}
(ii) \textit{diversity in the space of algorithms and features}, and
(iii) \textit{meta-learning for evaluating the cost and utility of detectors}.
REMIX provides two distinct ways for the analyst to consume its results: (i) a
partitioning of the detectors explored by REMIX into \textit{perspectives}
through low-rank non-negative matrix factorization; each perspective
can be \textit{easily visualized as an intuitive heatmap of experiments versus outliers}, and
(ii) an \textit{ensembled set of outliers} which combines outlier scores from all detectors.
We demonstrate the benefits of REMIX through extensive empirical validation on real-world data.
\end{abstract}

\section{Introduction}
Outlier detection is the identication of points in a dataset that do not conform to the norm.
This is a critical task in data analysis and is widely used in many applications
such as financial fraud detection, Internet traffic monitoring, and cyber security ~\cite{aggarwal2013outlier}.
Outlier detection is highly sensitive to the choice of the detection algorithm
and the feature subspace used by the algorithm ~\cite{aggarwal2013outlier, aggarwal2015theoretical, Zimek:2014:EUO:2594473.2594476}.
Further, outlier detection is often performed on high dimensional data
in an unsupervised manner without data labels; distinct sets of outliers discovered through different algorithmic choices could
reveal complementary insights about
the application domain. Thus, unsupervised outlier detection is a data analysis task
which inherently requires a principled exploration of the diverse algorithmic choices that are available.

Recent advances in interactive data exploration ~\cite{Dimitriadou:2014:EAQ:2588555.2610523,
SemanticWindows, Siddiqui:2016:EDE:3025111.3025126, conf/bigdata/WasayAI15} show much promise towards automated discovery of
advanced statistical insights from complex datasets while minimizing the burden of exploration for the analyst.
We study unsupervised outlier detection in
such an \textit{interactive setting} and consider three practical design requirements:
\textbf{(1) Automated exploration:} the system should automatically enumerate, assess, select and execute a diverse set of
outlier detectors; the exploration strategy should guarantee coverage in the space of features and algorithm parameters.
\textbf{(2) Predictable response time:} the system should conduct its exploration within a specified time limit.
This implies an exploration strategy that is sensitive to the execution time (cost)
of the detectors.
\textbf{(3) Visual interpretability:} the system should enable the user to easily navigate the results of automated
exploration by grouping together detectors that are similar in terms of the data points they identify as outliers.



\begin{figure}[th]
\centering
\subfigure[Perspective 1]{\label{perspective1}\includegraphics[width=88mm]{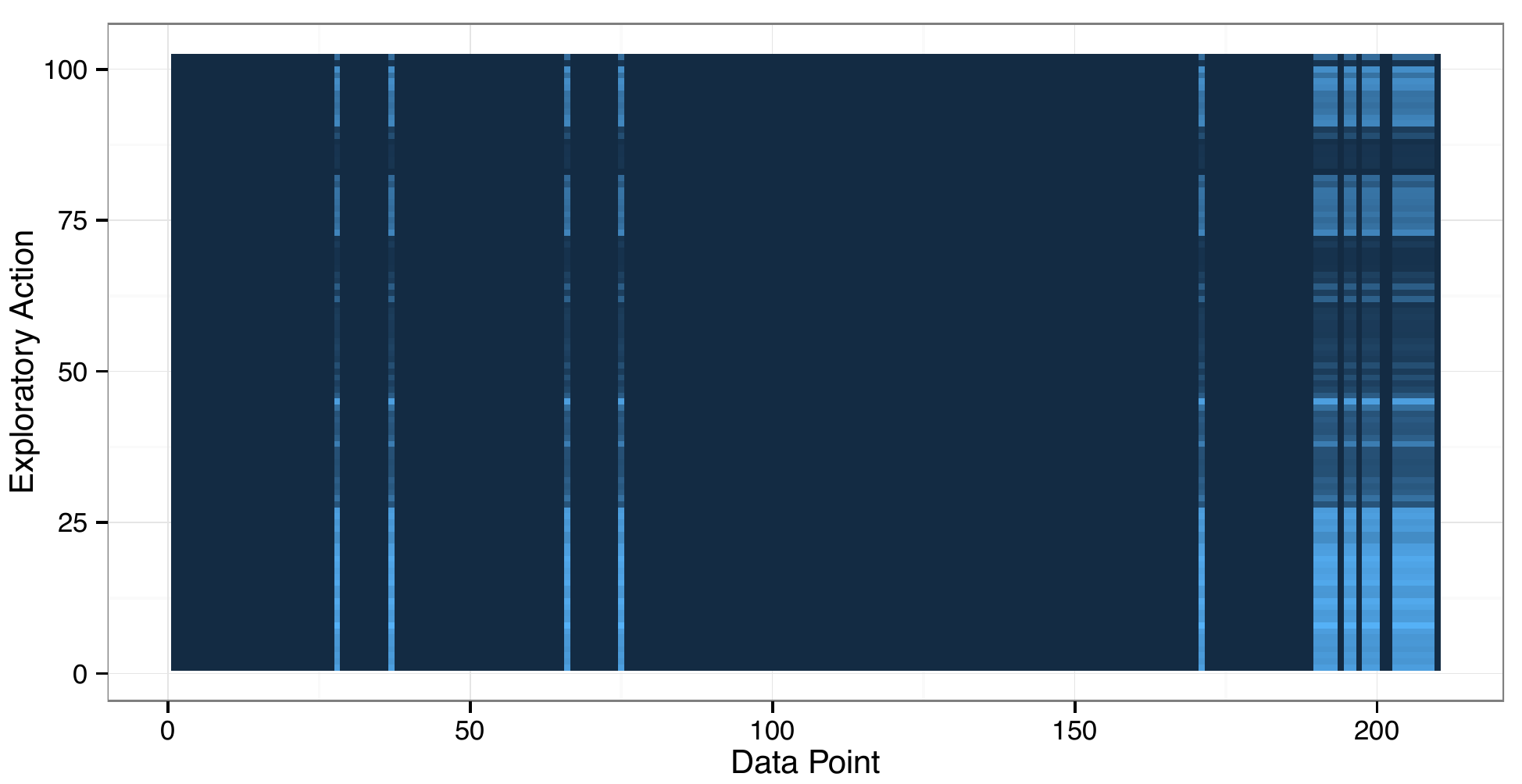}}
\subfigure[Perspective 2]{\label{perspective2}\includegraphics[width=88mm]{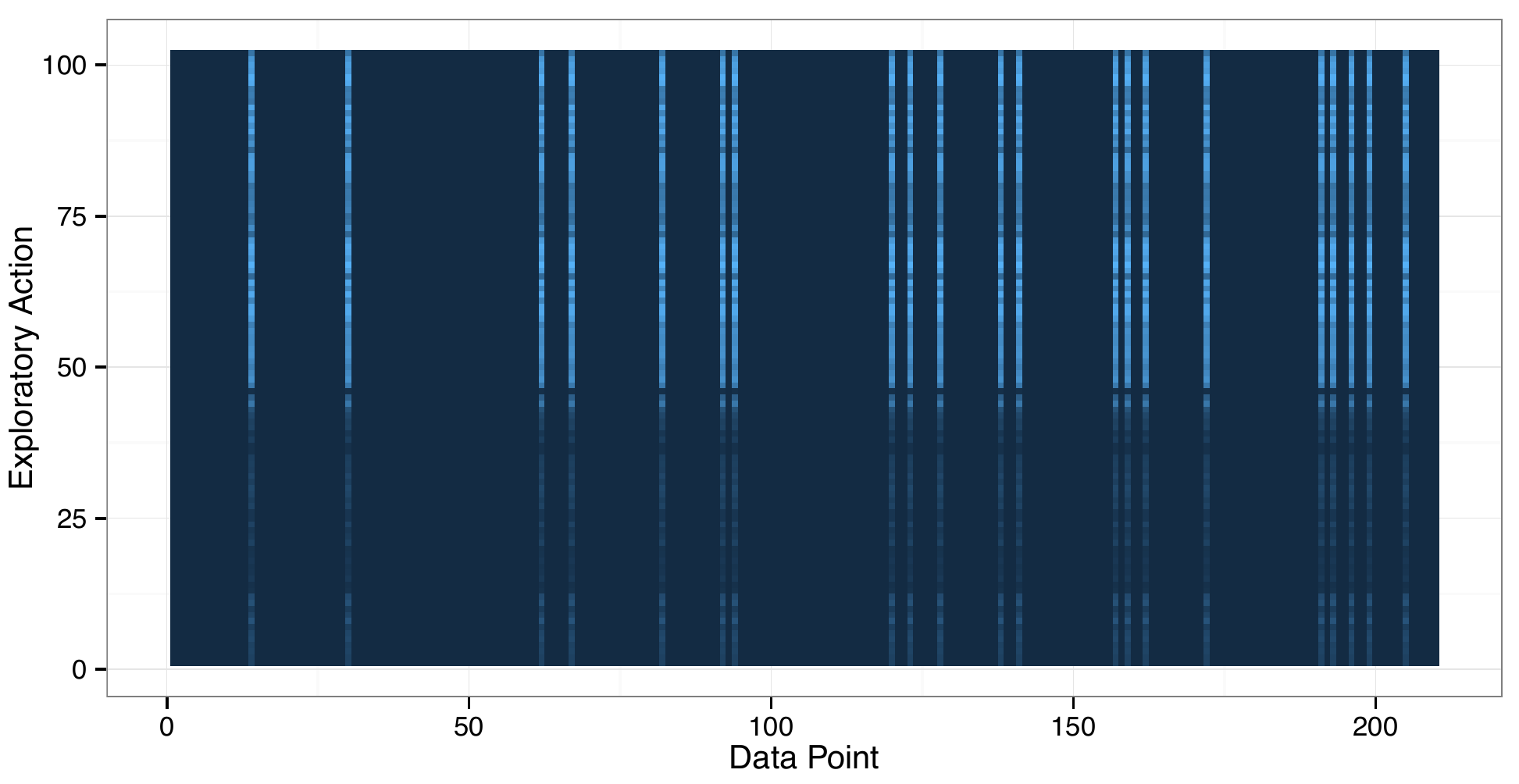}}
\caption{Factorization of an outlier matrix into two perspectives. Each perspective is a
heatmapped matrix whose columns are data points and rows are detectors. The intensity
of a cell $(s,p)$ in a perspective corresponds to the extent to which detector $s$ identifies
point $p$ as an outlier. Each perspective clearly identifies a distinct set of outliers.}
\label{fig:perspectives}
\end{figure}

\subsection{Key Contributions}
We present REMIX, a \textit{modular} framework for automated outlier exploration and the first
to address the outlier detection problem in an interactive setting.
To the best of our knowledge, none of the existing outlier detection systems support interactivity in the manner outlined above --
in particular, we are not aware of any system which automatically explores the diverse algorithmic choices in outlier detection
within a given time limit. The following are the key contributions of our work.

\subsubsection{MIP based Automated Exploration} REMIX systematically enumerates candidate outlier detectors and formulates the
exploration problem as a mixed integer program (MIP). The solution to this MIP yields a subset of candidates which are
executed by REMIX. The MIP maximizes a novel \textit{aggregate utility measure} which trades off between the
total utility of the top-$k$ vs all selected candidates; the MIP
also enforces  (i) an upper limit on the total cost (budget)
of the selected candidates, and (ii) a set of \textit{diversity constraints} which ensure that each
detection algorithm and certain prioritized feature subspaces get at least a certain minimum share of the exploration budget.

\subsubsection{Meta-Learning for Cost and Utility} The REMIX MIP requires an estimate of cost and utility
for each candidate detector. In order to estimate cost, REMIX trains a meta-learning model for each
algorithm which uses the number of data points, size of the feature subspace,
and various product terms derived from them as meta-features. It is significantly harder to estimate or even define utility.
REMIX handles this by defining the utility of a detector as a proxy for its accuracy on the given data.
REMIX estimates this by training a meta-learning model for each algorithm that uses various statistical
descriptors of a detector's feature subspace as meta-features; this model is trained on a corpus of
\textit{outlier detection datasets that are labeled} by domain experts~\cite{datasets}.

\subsubsection{Perspective Factorization} The diversity of feature subspaces and algorithms explored by REMIX
will result in different detectors marking a distinct set of data points as outliers. REMIX provides a succinct way for the
data analyst to visualize these results as heatmaps. Consider the outlier matrix $\Delta$
where $\Delta_{s, p}$ is the normalized ($[0, 1]$-ranging) outlier score assigned by detector $s$ to data point $p$.
REMIX uses a low rank non-negative matrix factorization
scheme to bi-cluster $\Delta$ into a small user-specified number of \textit{perspectives} such that
there is consensus among the detectors within a perspective. The idea of outlier perspectives is
\textit{a generalization of the idea of outlier ensembles}. A notable special case occurs when the number of
perspectives equals one: here, the results from all the detectors are \textit{ensembled}
into a single set of outliers.

REMIX can be used in two modes. \textbf{(i) Simple:} An analyst can gain rapid visual understanding of the
outlier space by providing two simple inputs: an \textit{exploration budget} and the
\textit{number of perspectives}. This yields bi-clustered heatmaps of outliers that are easily interpretable
as in Figure \ref{fig:perspectives}. \textbf{(ii) Advanced:} As discussed in Section \ref{sec:discussutility}, an
advanced user can also \textit{re-configure} specific modules within REMIX such as utility estimation or the
enumeration of prioritized feature subspaces. Such re-configuration would steer the REMIX MIP towards alternate
optimization goals while still guaranteeing exploration within a given budget and
providing factorized heatmap visualizations of the diverse outlier sets.

The rest of the paper is organized as follows. We survey related work in Section \ref{sec:related},
and provide background definitions and an overview in Section \ref{sec:preliminaries}.
In Sections \ref{sec:enumeration} -- \ref{sec:perspectives}, we present the details of
feature subspace and candidate detector enumeration, cost and utility
estimation, MIP for automated exploration, and perspective factorization.
We present an evaluation of REMIX on real-world datasets in Section \ref{sec:evaluation} and
conclude in Section \ref{sec:conclusion}.

\section{Related Work}\label{sec:related}
\textbf{Outlier ensembles}~\cite{aggarwal2013outlier,lazarevic2005feature, rayana2015less, aggarwal2015theoretical}
combine multiple outlier results to obtain a more robust set of outliers. Model centered ensembles
combine results from different base detectors while data-centered ensembles explore different
horizontal or vertical samples of the dataset and combine their results~\cite{aggarwal2013outlier}.
Research in the area of subspace mining ~\cite{lazarevic2005feature, keller2012hics, conf/cikm/KellerMWB13,
conf/sdm/BohmKMNV13, muller2011statistical} focuses on exploring a \textit{diverse family of feature subspaces}
which are interesting in terms of their ability to reveal outliers.
The work in \cite{alex_2015} introduces a new ensemble model
that uses detector explanations to choose base detectors selectively
and remain robust to errors in those detectors.

REMIX is related to ensemble outlier detection since setting the number of perspectives to one
in REMIX leads to ensembling of results from the base detectors. However, REMIX
has some notable distinctions: \textit{the idea of perspectives in REMIX generalizes the notion of ensembles};
setting the number of perspectives to a number greater than one is possible in REMIX and results
in complementary views of the outlier space which can be visualized as heatmaps; further, REMIX
seeks to guarantee coverage not just in the space of features or feature subspaces, but also available
algorithms -- \textit{subject to a budget constraint on the total time available for exploration}. None of the
existing approaches in literature provide this guarantee on the exploration time.


\textbf{Multiview outlier detection}~\cite{das2010multiple, muller2012outlier, marcos2013clustering, gao2011spectral} deals with a setting where the input consists of multiple
datasets (views), and each view provides a distinct set of features for characterizing the
objects in the domain. Algorithms in this setting aim to detect objects that are outliers in each view
or objects that are normal but show inconsistent class or clustering characteristics across different views.
REMIX is related to multiview outlier detection through non-negative matrix factorization (NMF) which is often used
here for creating a combined outlier score of objects. In REMIX, NMF is
used not just for ensembling detector scores but also for factorizing detector results into multiple heatmap
visualizations.

\textbf{Automated exploration} is a growing trend in the world of commercial data science systems~\cite{datarobot, skytree}
as well as machine learning and statistical research~\cite{DBLP:conf/aaai/BiemBFKMNPRRSST15, DBLP:conf/aaai/SabharwalST16,
khurana2016cognito, Lloyd2014-ABCD, DuvLloGroetal13, grosse2012exploiting}. While ~\cite{grosse2012exploiting} focuses on
model selection, \cite{Lloyd2014-ABCD, DuvLloGroetal13} focus on exploration for non-parametric regression models,
~\cite{datarobot, skytree, DBLP:conf/aaai/BiemBFKMNPRRSST15, DBLP:conf/aaai/SabharwalST16} deal with algorithm
exploration for classification models, while ~\cite{khurana2016cognito} deals with automated feature generation for
classification and regression. \textbf{Budgeted allocations and recommendations} have also been studied in
the context of problems other than outlier analysis ~\cite{li2011personalized, lu2011budgeted}.

\section{Preliminaries}
\label{sec:preliminaries}
\subsection{Baseline Algorithms}
Our implementation of REMIX uses the following set
$\mathcal{A}$ of five baseline outlier detection algorithms that are well-known.
\textbf{1) Local outlier factor:} LOF~\cite{breunig2000lof} finds outliers by measuring the local deviation
of a point from its neighbors. \textbf{2) Mahalanobis distance:} MD~\cite{hodge2004survey} detects
outliers by computing the Mahalanobis distance from a point and the center of the entire dataset as outlier score.
\textbf{3) Angle-based outlier detection:} ABOD~\cite{kriegel2008angle} identifies outliers by considering the
variances of the angles between the difference vectors of data points, which is more robust that distance in
high-dimensional space. ABOD is robust yet time-consuming. \textbf{4) Feature-bagging outlier detection:}
FBOD~\cite{lazarevic2005feature} is an ensemble method, which is based on the results of local outlier factor (LOF).
During each iteration, a random feature subspace is selected.
LOF then is applied to calculate the LOF scores based on the selected data subset.
The final score of FBOD is the cumulative sum of each iteration. \textbf{5) Subspace outlier detection:}
SOD~\cite{kriegel2009outlier}: SOD aims to detect outliers in varying subspaces of a high dimensional feature space.
Specifically, for each point in the dataset, SOD explores the axis-parallel subspace spanned by its neighbors and
determines how much the point deviates from the neighbors in this subspace.

\subsection{Interactive Outlier Exploration}
A dataset in REMIX is a real-valued matrix $A$ with $m$ columns and $n$ rows. A feature
subspace is a subset of columns in $A$. An outlier detector $D_{a, f}$ is simply
a combination of an algorithm $a \in \mathcal{A}$ and a feature subspace $f$ of $A$. The cost and utility values
$c_{a, f}$ and $u_{a, f}$ are positive values associated with $D_{a, f}$ which are intended to be estimates of the
execution cost and accuracy of $D_{a,f}$ respectively. REMIX enumerates candidate outlier detectors based on a
family of prioritized feature subspaces $\mathcal{F}_p$ and a family of randomly constructed feature subspaces
$\mathcal{F}_r$.

The \textit{interactive outlier exploration problem} is a budgeted optimization problem which selects a subset
of detectors with a \textbf{maximization objective} that is a linear combination of two quantities:
(i) the total utility of all the selected detectors, and (ii) the total utility of the top-$k$ selected detectors,
\textbf{subject to} the following budget and diversity constraints:
(i) the total cost the selected detectors does not exceed the budget $T_{total}$,
(ii) each algorithm gets a guaranteed share of the exploration budget, and
(iii) each prioritized feature gets a guaranteed share of the exploration budget.

\section{The REMIX Framework}
\begin{figure}[tb]
\centering
\includegraphics[width=75mm]{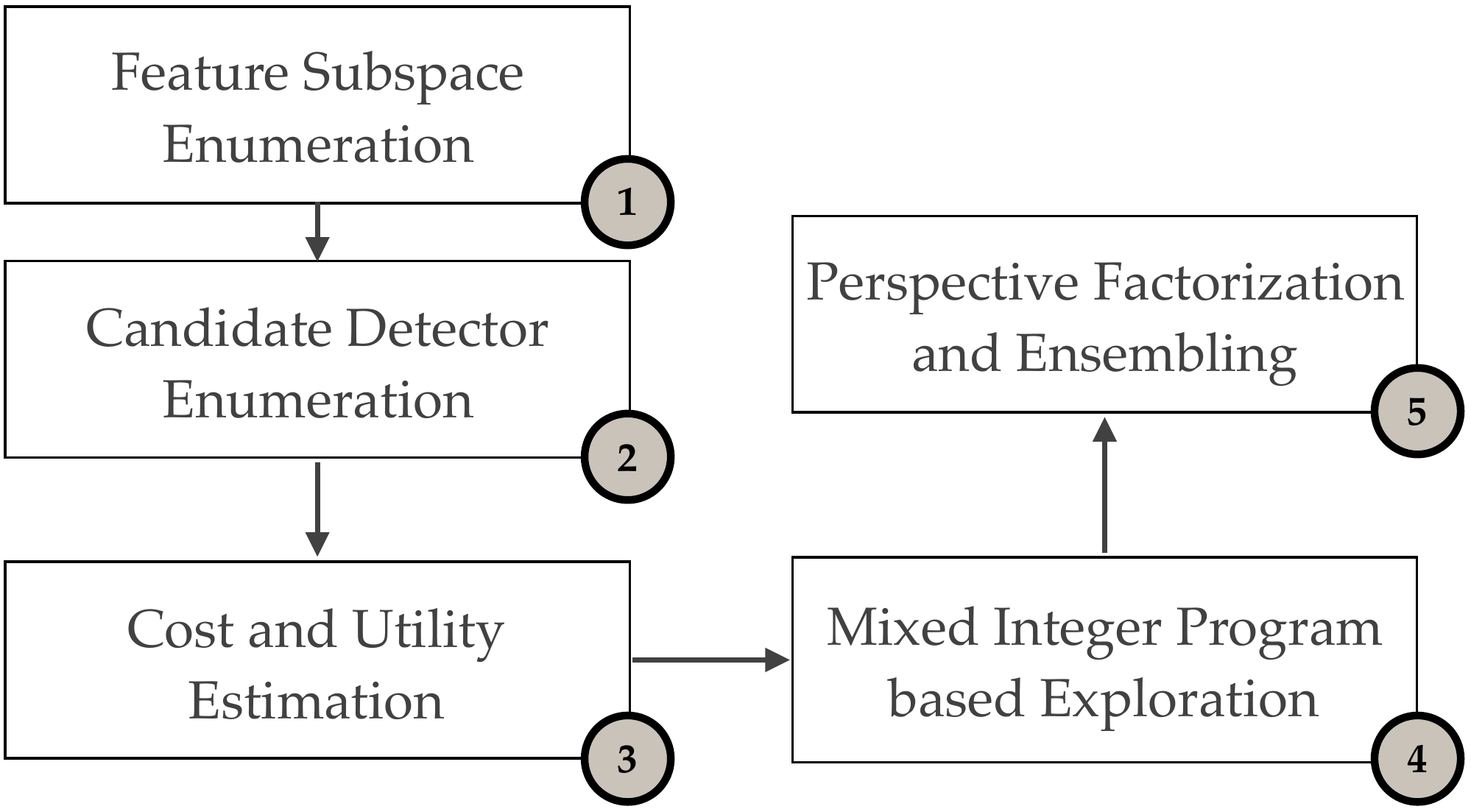}
\caption{REMIX Components: REMIX starts by enumerating multiple feature subspaces from the
given dataset, which in turn is used to enumerate candidate detectors. Next, REMIX evaluates the
cost and utility of all the enumerated detectors (the meta-learning components for training the cost
and utility models is not shown in this figure). The cost and utility values are used as part of a mixed
integer program (MIP) which selects a subset of candidates for execution based on a utility maximization
objective and budget and diversity constraints. The outlier results from the detectors executed by REMIX
are factorized into perspectives, and also ensembled into a single set of results.}
\label{fig:framework}
\end{figure}

We now describe the five components of REMIX shown in Figure \ref{fig:framework}
starting with feature subspace enumeration.

\subsection{Feature Subspace and Candidate Detector Enumeration}
\label{sec:enumeration}
\begin{algorithm}
  \caption{Feature Subspace Enumeration}\label{algo:subspace}
  \begin{algorithmic}[1] 
    \Require Data Matrix $A$
    \Ensure Feature subspace families $\mathcal{F}_p$ and $\mathcal{F}_r$
    \State $F_{nr} = \bigcup\limits_{j=1}^{m} A_{*, j}$ \Comment{Initialize non-redundant feature bag}
    \While{$\vert F_{nr} \vert \geq 2$} \Comment{Bag has at least 2 features}
      \State $\forall A_{*,j} \in F_{nr}, \hat{\sigma}_j \gets \frac{\sum_{(i \neq j) \wedge (A_{*, i} \in F_{nr})} \sigma_{i,j}}{\vert F_{nr} \vert - 1}$ \Comment{Mean correlation}
      \State $(p, q) \gets \arg\max_{(i \neq j) \wedge (A_{*, i} \in F_{nr}) \wedge (A_{*,j} \in F_{nr})}{\sigma_{i,j}}$
      \Comment{Max}
      \If{$\sigma_{p, q} \geq \alpha$} \Comment{High max correlation?}
        \If{$\hat{\sigma}_p \geq \hat{\sigma}_q$} \Comment{Greater average correlation?}
        \State $F_{nr} \gets F_{nr} \setminus \{A_{*, p}\}$
        \Comment{Drop $A_{*, p}$}
        \Else
          \State $F_{nr} \gets F_{nr} \setminus \{A_{*, q}\}$
          \Comment{Drop $A_{*, q}$}
        \EndIf
          \Else \textbf{ break} \Comment{Break out of while loop}
      \EndIf
    \EndWhile

    \State $\mathcal{F}_p = \Phi$ \Comment{Initialize to null set}
    \For{$\ell \gets 1, \vert F_{nr} \vert$}
      \State $S_\ell \gets$ Top-$\ell$ features in $F_{nr}$ ranked by their Laplacian scores
      \State $\mathcal{F}_p \gets \mathcal{F}_p \cup \{S_\ell\}$ \Comment{Add prioritized subspace}
    \EndFor

    \State $\mathcal{F}_r = \Phi$ \Comment{Initialize to null set}
    \For{$i \gets 1, \Gamma$}
      \State $T_i \gets$ Random subspace with each feature sampled independently at random
      without replacement from $F_{nr}$ with probability $\frac{1}{2}$
      \State $\mathcal{F}_r \gets \mathcal{F}_r \cup \{T_i\}$ \Comment{Add random subspace}
    \EndFor
  \end{algorithmic}
\end{algorithm}

Algorithm \ref{algo:subspace} describes feature subspace enumeration and has three parts: (i) creating a
non-redundant feature bag $F_{nr}$ (lines 1 -- 13), (ii) creating
a prioritized family of subspaces $\mathcal{F}_p$ (lines 14-18) using a feature ranking approach and (iii) creating
 a randomized family of
subspaces $\mathcal{F}_r$ (lines 19 -- 23) used for maximizing coverage and diversity during exploration. Redundant features
$\not\in F_{nr}$ are not part of subspaces in $\mathcal{F}_p$ or $\mathcal{F}_r$.

Algorithm \ref{algo:subspace} begins by initializing $F_{nr}$ to all features in $A$ (line 1). It then iteratively
looks for a member in $F_{nr}$ which can be considered redundant and hence dropped from $F_{nr}$. In order for a
feature $A_{*, p} \in F_{nr}$ to be considered redundant, it needs to (i) be in a maximally correlated feature pair
$\{A_{*, p}, A_{*, q}\} \subseteq F_{nr}$ (line 4), (ii) the correlation coefficient $\sigma_{p, q}$ must be above the REMIX's
redundancy threshold $\alpha$ (line 5), and (iii) of the two features in the pair, $A_{*, p}$ must have a mean correlation
with other features in $F_{nr}$ that $\geq$ the mean correlation of $A_{*, q}$ (line 7). We
set the default value of $\alpha$ in REMIX to 0.9 based on our experimental evaluation. It is easy to see that at the
end of line 13, Algorithm \ref{algo:subspace} yields a non-redundant feature bag $F_{nr}$ with the following property.

\begin{observation}
\label{observation:nonredundant}
For any pair of features $\{A_{*, p}, A_{*, q}\} \subseteq F_{nr}$, $\sigma_{p, q} < \alpha$. For any feature $A_{*, r} \not\in F_{nr}$,
$\exists A_{*, s} \in F_{nr}, \sigma_{r, s} \geq \alpha$.
\end{observation}

The family of randomized feature subspaces $\mathcal{F}_r$ is created for the purpose of guaranteeing feature coverage
and diversity during exploration.
In particular, we select $\Gamma$ subspaces, where each subspace consists of
features selected independently at random without replacement with probability $\frac{1}{2}$ from $F_{nr}$.
We set the default value of $\Gamma$ in REMIX to $|\mathcal{F}_p|/2$ to balance the size of the two families.

The family of prioritized feature subspaces $\mathcal{F}_p$ is created as follows. The features in $F_{nr}$
are first sorted according to their Laplacian scores~\cite{he2005laplacian}. We add $\vert F_{nr} \vert$ subspaces to $\mathcal{F}_p$,
where the $\ell^{th}$ subspace is the set of top-$\ell$ features in $F_{nr}$ ranked by their Laplacian scores.
We now provide a brief justification for our use of the Laplacian score. Due to lack of space, refer the reader to
~\cite{he2005laplacian} for the exact details of the Laplacian computation. The Laplacian score is computed
as a way to reflect a feature's ability to preserve locality. In particular, consider the projection $A'$ of all
points in $A$ onto the subspace $F_{nr}$; now, consider the $r$-nearest neighborhood of points in $A'$. Features
that respect this $r$-nearest neighborhood provide a better separation of the inlier class from outlier class
within the data. We note that our experimental evaluation in Section \ref{sec:evaluation} consistently demonstrates
improved outlier detection accuracy with $\mathcal{F}_p$ and $\mathcal{F}_r$ as opposed to purely $\mathcal{F}_r$.
We also discuss potential alternatives for prioritized feature subspace construction in Section \ref{sec:discussutility}.

We enumerate candidate detectors by a cartesian product of $\mathcal{F}_p \cup \mathcal{F}_r$ and
the set of baseline detection algorithms $\mathcal{A}$.

\subsection{Cost Estimation}
\label{sec:costestimation}
The cost of a candidate detector can be modeled as a function of its algorithm as well as the size of its
feature subspace $f$ ($n$ and $\vert f \vert$). The runtime complexity of algorithms in the `big-O' notation provides an asymptotic
relationship between cost and input size; however, in REMIX, we seek a more refined model which accounts
for lower order terms and the constants hidden by `big-O'. We do this by
training a multivariate linear regression model for \textit{each algorithm}. Specifically, consider the following polynomial:
$ (1 + \vert f \vert + n + \log{\vert f \vert} + \log{n})^3 - 1$.
There are ${7 \choose 3} - 1$ distinct terms in the expansion of this polynomial which can be derived exactly given
a feature subspace. The terms\footnote{The exponent 3 suffices to model the cost of
most known outlier detection algorithms} form the explanatory variables while cost
is the dependent variable in the linear regression model.

\subsection{Utility Estimation}
\label{sec:utilityestimation}
Given a detector $D_{a, f}$, the utility estimation algorithm (Algorithm \ref{algo:utilityestimation})
first normalizes the features in $f$, and computes a variety of feature-level statistics for each feature $\psi \in f$. These
statistics include the Laplacian score which is a useful measure for unsupervised feature selection
(Section \ref{sec:enumeration}), the standard deviation, skewness which measures
the assymetry of the feature distribution, kurtosis which measures the extent to which the feature distribution is heavy-tailed,
and entropy which is a measure of the information content in the feature. We note that these computations are done once for
each feature in $F_{nr}$ and is reused for any given any detector. Next, the algorithm computes the meta-feature vector $MFV(f)$ of
\textit{feature-subspace-level} statistics by combining feature-level statistics. For instance, consider the Laplacian scores of all
the features in $f$; the mean, median, median~absolute~deviation, min, max, and standard~deviation of all the Laplacian scores
provide $6$ of the $30$ distinct components of $MFV(f)$ in this step. Finally, it uses an algorithm
specific utility model $U_{a}(MFV(f))$ to estimate the utility $u_{a, f}$ of $D_{a, f}$. REMIX trains five distinct linear regression models
$U_{LOF}$, $U_{MD}$, $U_{ABOD}$, $U_{LOF}$, and $U_{SOD}$ corresponding to each algorithm. These models are trained based on
distinct \textit{expert labeled} datasets from the outlier dataset repository~\cite{datasets}.
The explanatory variables for this linear regression are the feature-subspace-level statistics described in Algorithm \ref{algo:utilityestimation}. The dependent variable is
the detection accuracy, measured as the fraction of the data points on which both the detector and the expert labeled
ground truth agree on the outlier characterization.

\begin{algorithm}
  \caption{Utility Estimation}\label{algo:utilityestimation}
  \begin{algorithmic}[1] 
    \Require A candidate detector $D_{a, f}$
    \Ensure The utility $u_{a, f}$ of the detector
    \State $\forall \psi \in f$, normalize $\psi$ \Comment{One time procedure applied to $F_{nr}$}
    \State $\forall \psi \in f$, extract the following $5$ feature-level statistics:
    Laplacian score, standard deviation, skewness, kurtosis, and entropy \Comment{One time procedure applied to $F_{nr}$}
    \State Extract the following $30$ feature-subspace-level statistics:
    mean, median, median~absolute~deviation, min, max, and standard~deviation for each of the $5$
    feature-level statistics extracted in Step 2. Let $MFV(f)$ contain these feature subspace-level statistics.
    \State Return $u_{a, f} = U_{a}(MFV(f))$, where $U_{a}$ is the utility estimation model learnt for algorithm $a$
  \end{algorithmic}
\end{algorithm}

\subsubsection{Discussion}
\label{sec:discussutility}
Estimating or even defining the utility of a detector is \textit{significantly harder} than estimating its cost. The
goal of utility estimation in REMIX is \textit{not} to learn a perfect model; rather,
the goal is merely to learn a utility model which can effectively steer the solution of the mixed integer program (MIP) used
by REMIX for exploration (Section \ref{sec:mip}). Our experiments in Section \ref{sec:evaluation} demonstrate that this is indeed the
case with REMIX. Further, REMIX is intended to be a flexible framework where alternative mechanisms for utility estimation
can be plugged in. For instance, consider Cumulative Mutual Information (CMI) metric and the Apriori-style
feature subspace search algorithm presented in ~\cite{conf/sdm/BohmKMNV13} for subspace outlier detection.
REMIX can use CMI as the detector utility, and this Apriori-style algorithm as an alternative for enumerating
prioritized feature subspaces. We chose the meta-learning approach in our implementation
since this approach is algorithm agnostic and hence can be generalized easily.

\subsection{MIP based Exploration}
\label{sec:mip}
REMIX executes only a subset of detectors enumerated by its detector enumeration component (Section \ref{sec:enumeration}).
This subset is determined by solving the following mixed integer program (MIP).

\begin{eqnarray}
&\max&\! \underbrace{\sum_{a \in \mathcal{A}} \sum_{f \in \mathcal{F}_p \cup \mathcal{F}_r} z_{a,f} u_{a,f} }_{\text{utility of top-k detectors}} +
\lambda \underbrace{\sum_{a \in \mathcal{A}} \sum_{f \in \mathcal{F}_p \cup \mathcal{F}_r} y_{a,f} u_{a,f}}_{\text{utility of all detectors}} \label{obj}
\\
&& \sum_{a \in \mathcal{A}} \sum_{f \in \mathcal{F}_p \cup \mathcal{F}_r} y_{a,f} c_{a,f} \leq T_{total} \label{constraint1}
\\
&& \forall a \in \mathcal{A}: \sum_{f \in \mathcal{F}_p \cup \mathcal{F}_r} y_{a,f} c_{a,f} \geq \frac{T_{total}}{2 \cdot \vert \mathcal{A} \vert} \label{constraint2}
\\
&& \forall f \in \mathcal{F}_p: \sum_{a \in \mathcal{A}} y_{a,f} c_{a,f} \geq \frac{T_{total}}{2 \cdot \vert \mathcal{F}_p \vert} \label{constraint3}
\\
&& \forall a \in \mathcal{A}, \forall f \in \mathcal{F}_p \cup \mathcal{F}_r: z_{a,f} \leq y_{a,f} \label{constraint5}
\\
&& \sum_{a \in \mathcal{A}} \sum_{f \in \mathcal{F}_p \cup \mathcal{F}_r} z_{a,f} \leq k \label{constraint4}
\\
&& \forall a \in \mathcal{A}, \forall f \in \mathcal{F}_p \cup \mathcal{F}_r: y_{i,j} \in \{0,1\} \label{constraint6}
\\
&& \forall a \in \mathcal{A}, \forall f \in \mathcal{F}_p \cup \mathcal{F}_r: z_{i,j} \in \{0,1\} \label{constraint7}
\end{eqnarray}

Given an algorithm $a$ and a feature subspace $f$, $y_{a,f}$ is the binary indicator variable in the MIP which determines
if the detector $D_{a, f}$ is chosen for execution in the MIP solution. Recall that $c_{a,f}$
denotes the estimated cost of $D_{a,f}$.
We observe the following.

\begin{observation}
  Constraint \ref{constraint1} guarantees that in any feasible solution to the MIP,
  the total cost of the selected detectors is $\leq T_{total}$.
\end{observation}

The value of the total exploration budget $T_{total}$ is provided by the analyst as part of her interaction
with REMIX.

\begin{observation}
  Constraint \ref{constraint2} guarantees that in any feasible solution to the MIP,
  each algorithm is explored for an estimated duration of time which is
  $\geq \frac{T_{total}}{2 \cdot \vert \mathcal{A} \vert}$.
\end{observation}

\begin{observation}
  Constraint \ref{constraint3} guarantees that in any feasible solution to the MIP,
  each prioritized feature is explored for an estimated duration of time which is
  $\geq \frac{T_{total}}{2 \cdot \vert \mathcal{F}_p \vert}$. This also implies that the exploration focuses
  at least half its total time on prioritized features.
\end{observation}

Consider the binary indicator variable $z_{a,f}$ corresponding to the detector $D_{a, f}$. Constraint $\ref{constraint5}$
ensures that in any feasible solution to the MIP, $z_{a,f}$ can be $1$ only if it is chosen in the solution (i.e., $y_{a,f} = 1$).
Constraint \ref{constraint4} ensures at most $k$ of the detectors chosen by the solution have their $z$-values set to $1$.
Now consider an \textit{optimal} solution to the MIP. Since utility values are non-negative for all detectors, the first
part of the objective function $\sum_{a \in \mathcal{A}} \sum_{f \in \mathcal{F}_p \cup \mathcal{F}_r} z_{a,f} u_{a,f}$
is maximized when exactly $k$ detectors have their $z$-values set to $1$ (and not fewer) \textbf{and} when the detectors whose
$z$-values are set to $1$ are the ones with the highest utilities amongst the selected detectors. This leads us to the
following guarantee.

\begin{theorem}
  The optimal solution to the MIP maximizes the sum of utilities of the top-$k$ detectors with highest utilities
  \textit{and} the total utility of all the selected detectors scaled by a factor $\lambda$.
\end{theorem}

We set $k = 10$ and $\lambda = 1$ in our implementation of REMIX which balances the utility of
the top-$10$ detectors vs the total utility of all the selected detectors.

\subsection{Perspective Factorization}
\label{sec:perspectives}
Each detector executed by REMIX provides an outlier score for each data point and the results
from different detectors could be potentially divergent. We now present a
factorization technique called \textbf{NMFE} (non-negative matrix factorization and ensembling) for
bi-clustering the detection results into a few succinct \textit{perspectives}. All detectors within a
perspective agree on how they characterize outliers although there could be disagreement across perspectives.

Let $\Delta_{s,p}$ be the outlier score assigned by detector $s$ for data point $p$ normalized across
data points to be in the range $[0, 1]$. Consider the matrix of outlier scores $\Delta \in [0, 1]^{t \times n}$,
where $t$ is the number of detectors executed by REMIX and $n$ is the number of data points.
We perform a rank-g non-negative matrix factorization of $\Delta \approx \Lambda\Omega^{\top}$,
where $\Lambda \in \mathbb{R}^{t \times g}$ and $\Omega \in \mathbb{R}^{n \times g}$,
by minimizing the Kullback-Leibler divergence between $\Delta$ and $\Lambda\Omega^\top$~\cite{Ding:2010:CSM:1687044.1687110}:
\begin{equation}
\min_{\Lambda, \Omega \geq 0} \sum_{s, p} \Delta_{s, p} log\frac{\Delta_{s, p}}{\Lambda_{s, *} \Omega_{p, *}} - \Delta_{s, p} + \Lambda_{s, *} \Omega_{p, *}
\label{equ:ensemble}
\end{equation}

The matrix $\Lambda\Omega^T$ by definition can be expressed as a sum of $g$ rank-1 matrices whose
rows and columns correspond to detectors and data points respectively.
In any of these rank-1 matrices, every row (column) is a scaled multiple of
any other non-zero row (column), and every entry is between $0$ and $1$. These properties make it
possible for the rank-1 matrices to be visualized as \textit{perspectives}, or heatmaps where the
intensity of a heatmap cell is the value of the corresponding entry in the perspective, as shown in
Figure \ref{fig:perspectives}. The number of perspectives $g$ is specified by the user as an input to REMIX.
Setting $g=1$ simply results in a direct averaging (ensembling) of all the
detector results into a single perspective.

\section{Experiments}
\label{sec:evaluation}
In this section, we present the evaluation of REMIX on real-world data.

\subsection{Data Collection}
\label{sec:data_description}

We chose 98 datasets for our study from the outlier dataset repository~\cite{datasets} of the Delft University of Technology.
This corpus contains outlier detection datasets labeled by domain experts. These datasets span varied domains
such as flowers, breast cancers, heart diseases, sonar signals, genetic diseases, arrhythmia abnormals, hepatitis,
diabetes, leukemia, vehicles, housing, and  satellite images.
Each dataset has benchmark outlier labels, with an entry 1 representing outliers and 0 representing normal points.

\begin{figure}[ht]
\centering
\subfigure[Distribution of feature sizes]{\label{fig:feature_size} \includegraphics[width=41mm]{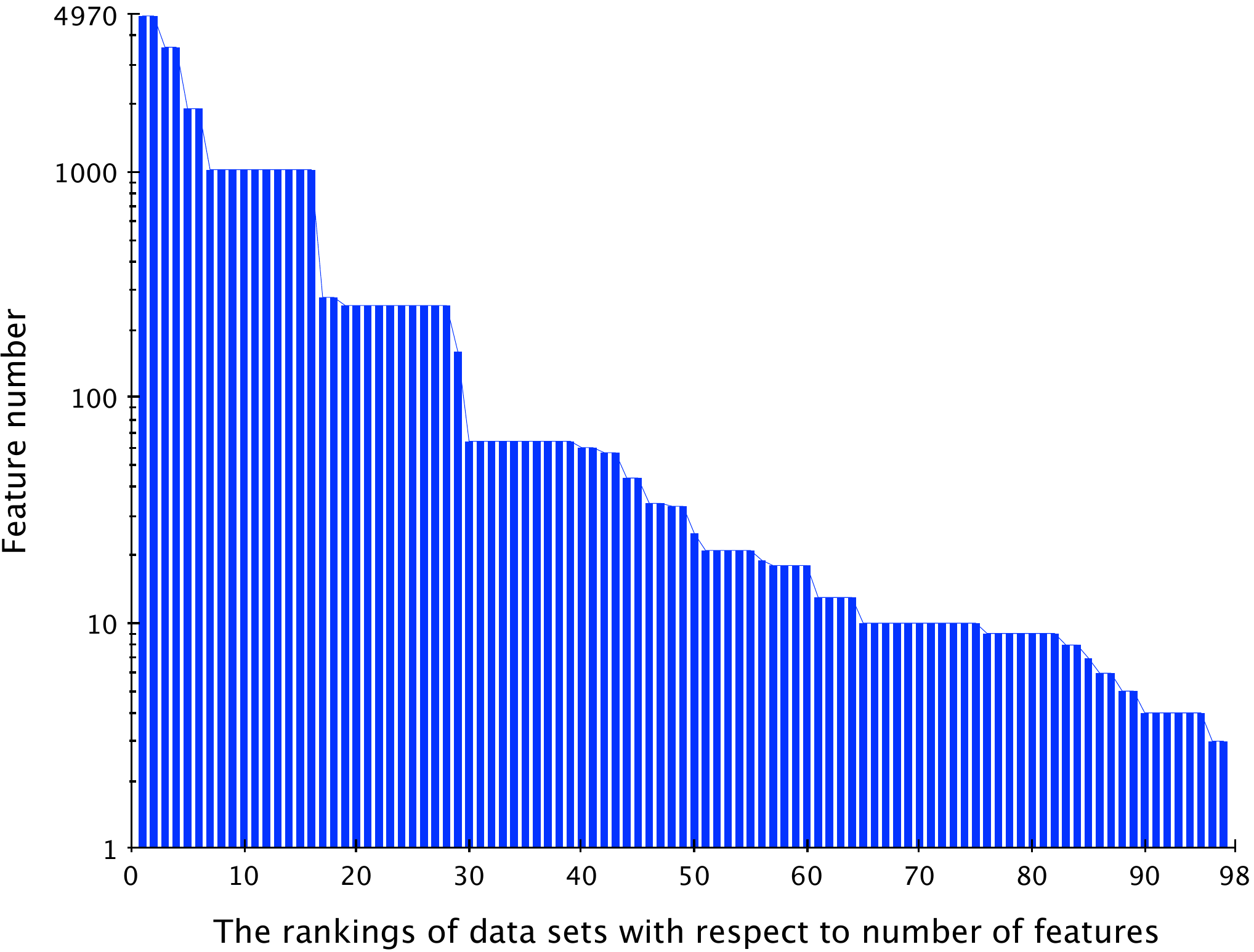}}
\subfigure[Distribution of outlier ratios]{\label{fig:outlier_ratio} \includegraphics[width=41mm]{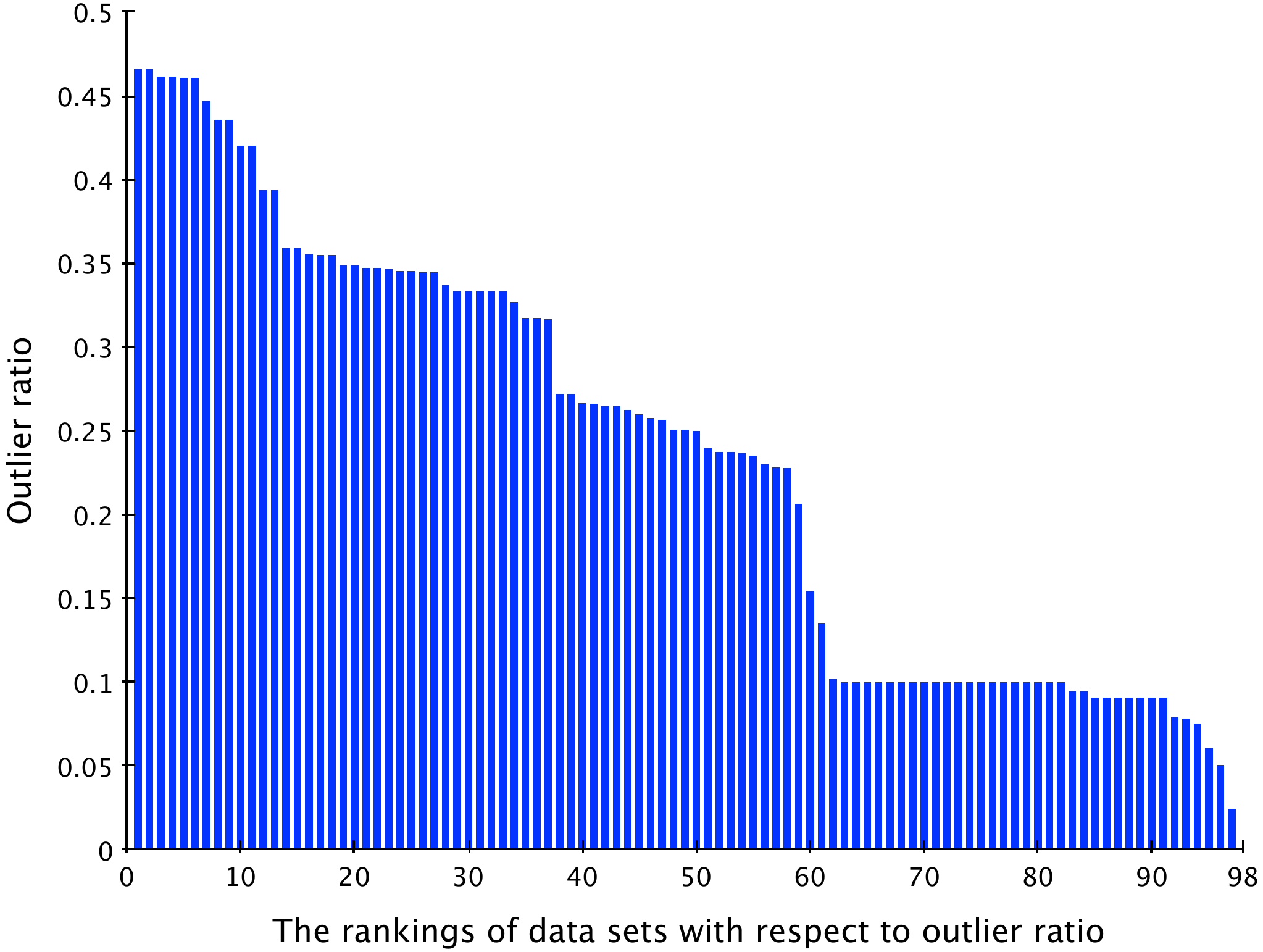}}
\caption{Statistics of the experimental datasets}
\label{fig:statistics}
\end{figure}

Figure~\ref{fig:statistics} illustrate some statistics of the 98 data sets. Specifically,
Figure \ref{fig:feature_size} shows the numbers of features for each dataset sorted in a descending order.
In this figure, we can observe that most of the datasets contains less than 100 features while only a small portion of these
datasets have more than 1000 features.
Figure \ref{fig:outlier_ratio} shows the outlier ratio to the total number of data points for each dataset sorted in
a descending order. In this figure, we can find that the outlier ratios of more than 50\% of the datasets are less than 23.2\%.

\subsection{Cost and Utility Estimation}
\begin{figure}[ht]
\centering
\subfigure[Cost Estimation]{\label{fig:cost} \includegraphics[width=41mm]{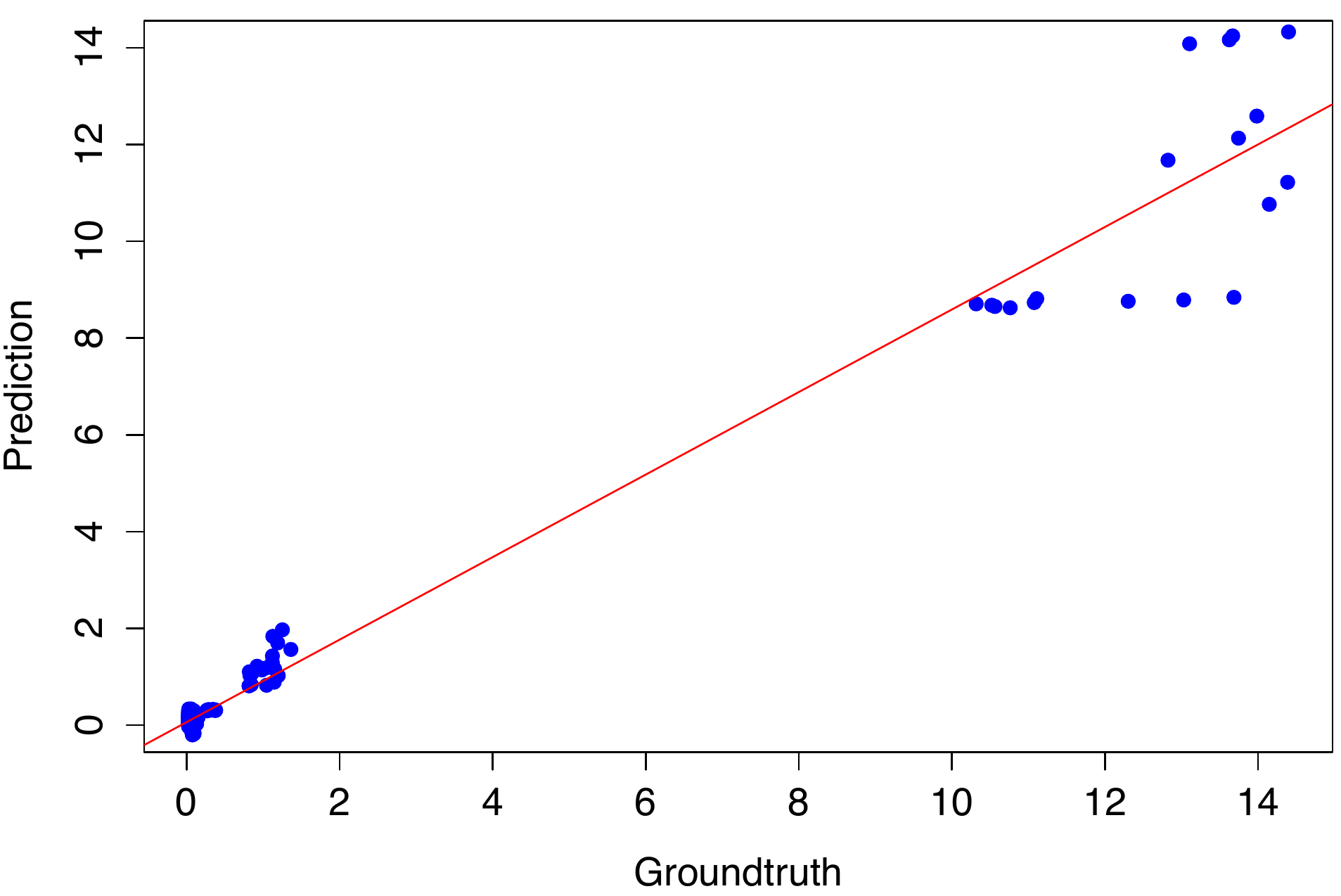}}
\subfigure[Utility Estimation]{\label{fig:utility} \includegraphics[width=41mm]{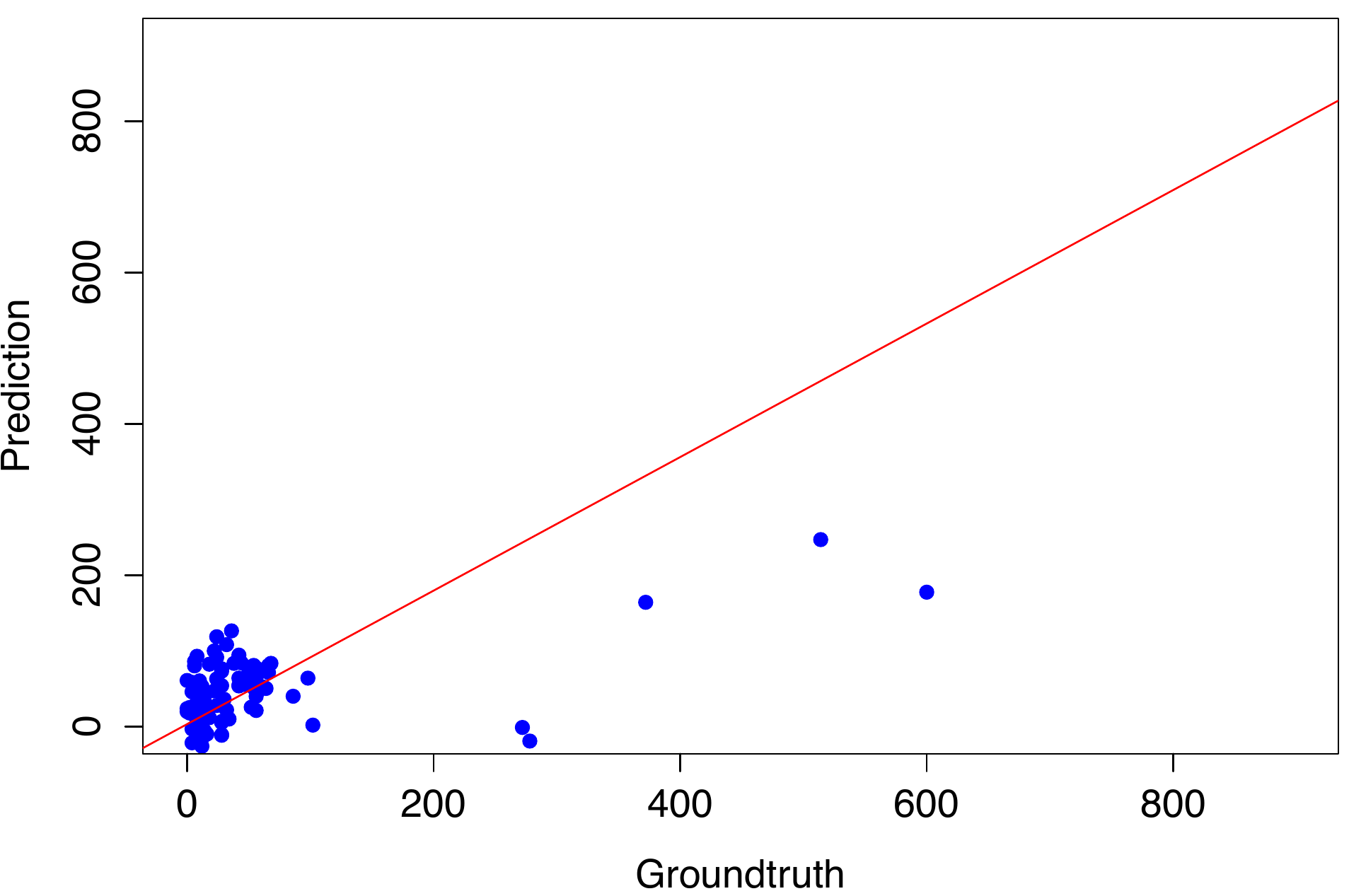}}
\caption{Cost and Utility Estimation}
\label{fig:estimation}
\end{figure}

Among the 98 datasets, we used 96 datasets to train and test the cost and utility estimation models
for each of the $5$ baseline algorithms in our REMIX implementation (with a 70\%-30\% split for train vs test).
Figures \ref{fig:cost} and \ref{fig:utility} present the performance of the cost and utility models for the
LOF algorithm from a specific run.

Recall that the utility $u_{a, f}$ estimates the fraction of the data points
on which the detector $d_{a, f}$ and the expert labels agree on the outlier characterization. Consider the Hamming distance
between the outlier bit vector (1 if outlier, 0 otherwise) created by the detector and the outlier bit vector
created by the expert labels. Clearly, this Hamming distance equals $n \cdot (1 - u_{a, f})$ where $n$ is the number
of data points. Figure \ref{fig:utility} plots the estimated Hamming distance on the $y$-axis and the actual
distance as observed by running the detector on the $x$-axis. The red lines in Figures \ref{fig:cost} and
\ref{fig:utility} represent ideal predictors. As expected, the cost estimator clearly performs better than
the utility estimator. However, as mentioned in Section \ref{sec:discussutility} our goal in utility estimation is
not a perfect predictive model but merely to steer the MIP towards better solutions. We demonstrate this to be the case
in our next set of experiments.

\subsection{Detection Accuracy and Cost}
\begin{figure*}[th]
\centering
\subfigure[Precision@N]{\includegraphics[width=43mm]{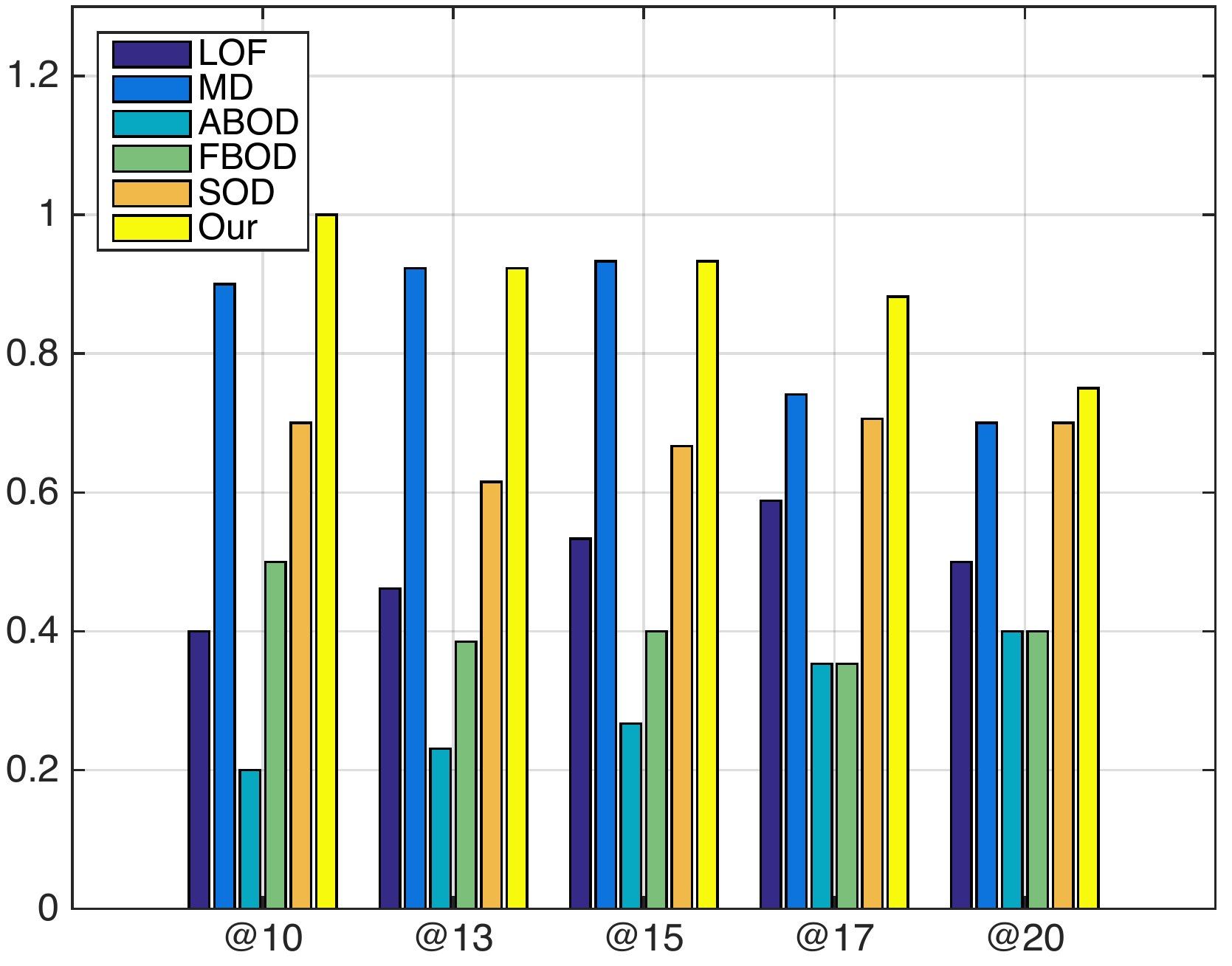}}
\subfigure[Recall@N]{\includegraphics[width=43mm]{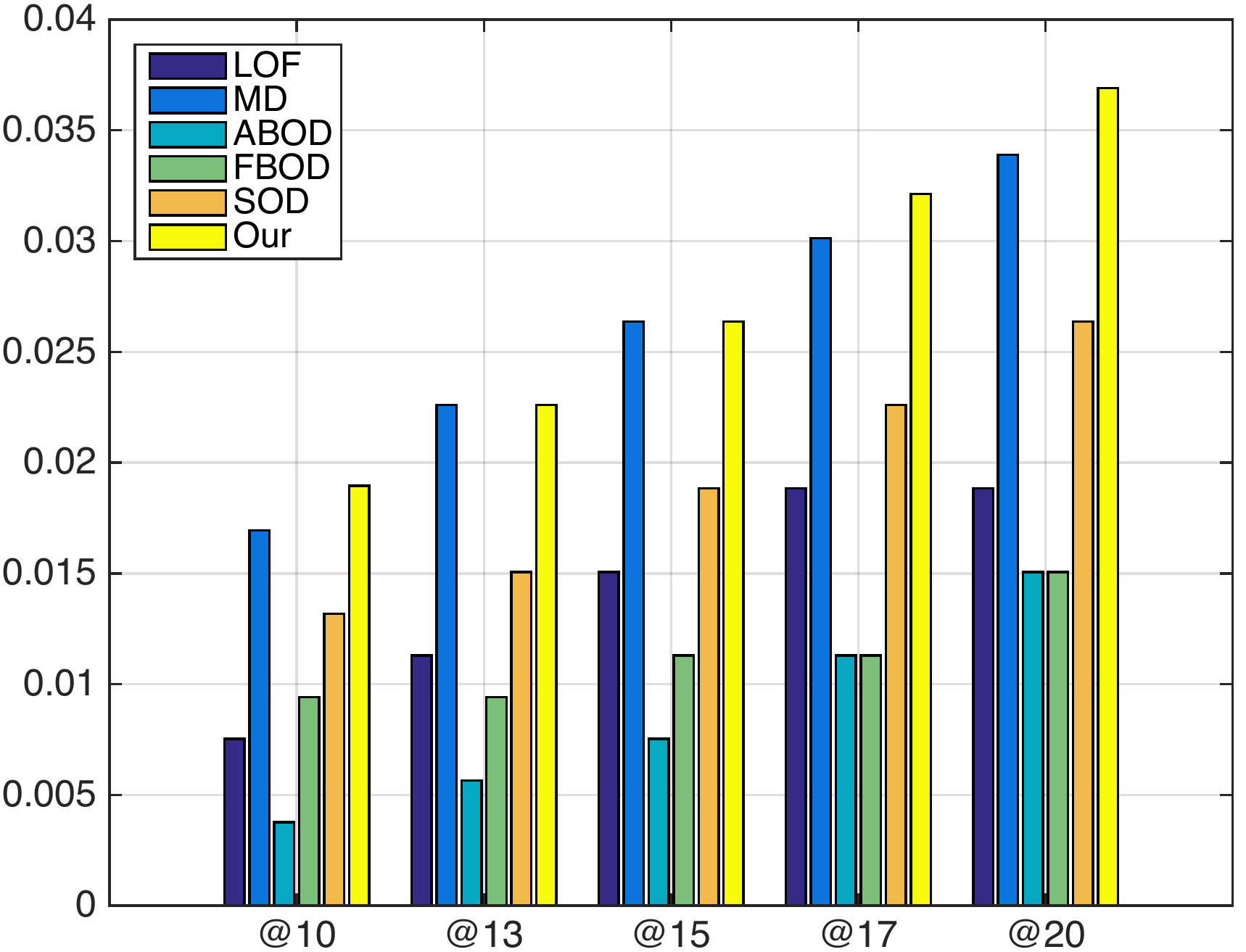}}
\subfigure[Fmeasure@N]{\includegraphics[width=43mm]{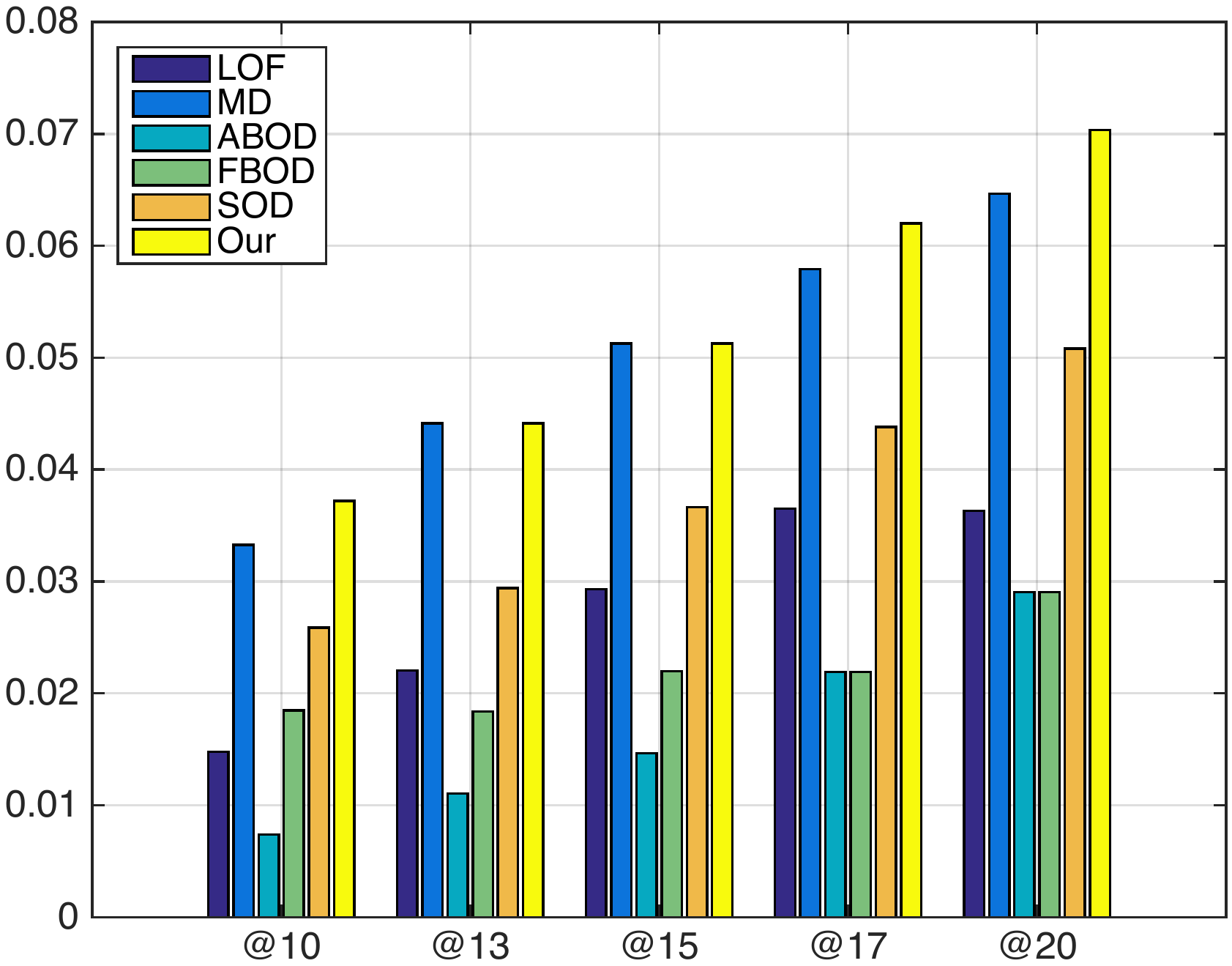}}
\vspace{-0.1 mm}
\caption{Comparison of REMIX with baseline algorithms on the cardiac arrhythmia dataset}
\label{overall_performance_oc541}
\end{figure*}

\begin{figure*}[h]
\centering
\subfigure[Precision@N]{\includegraphics[width=43mm]{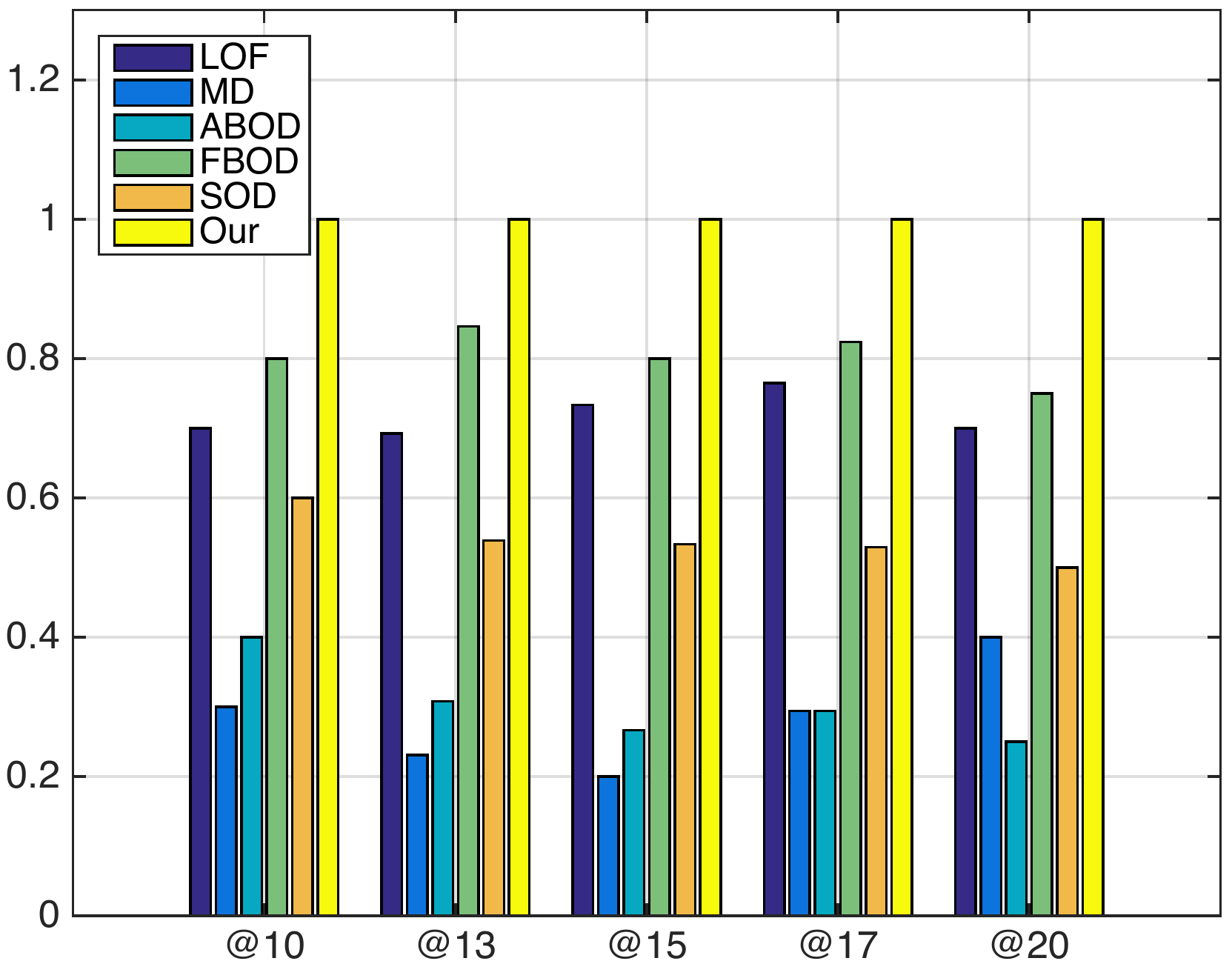}}
\subfigure[Recall@N]{\includegraphics[width=43mm]{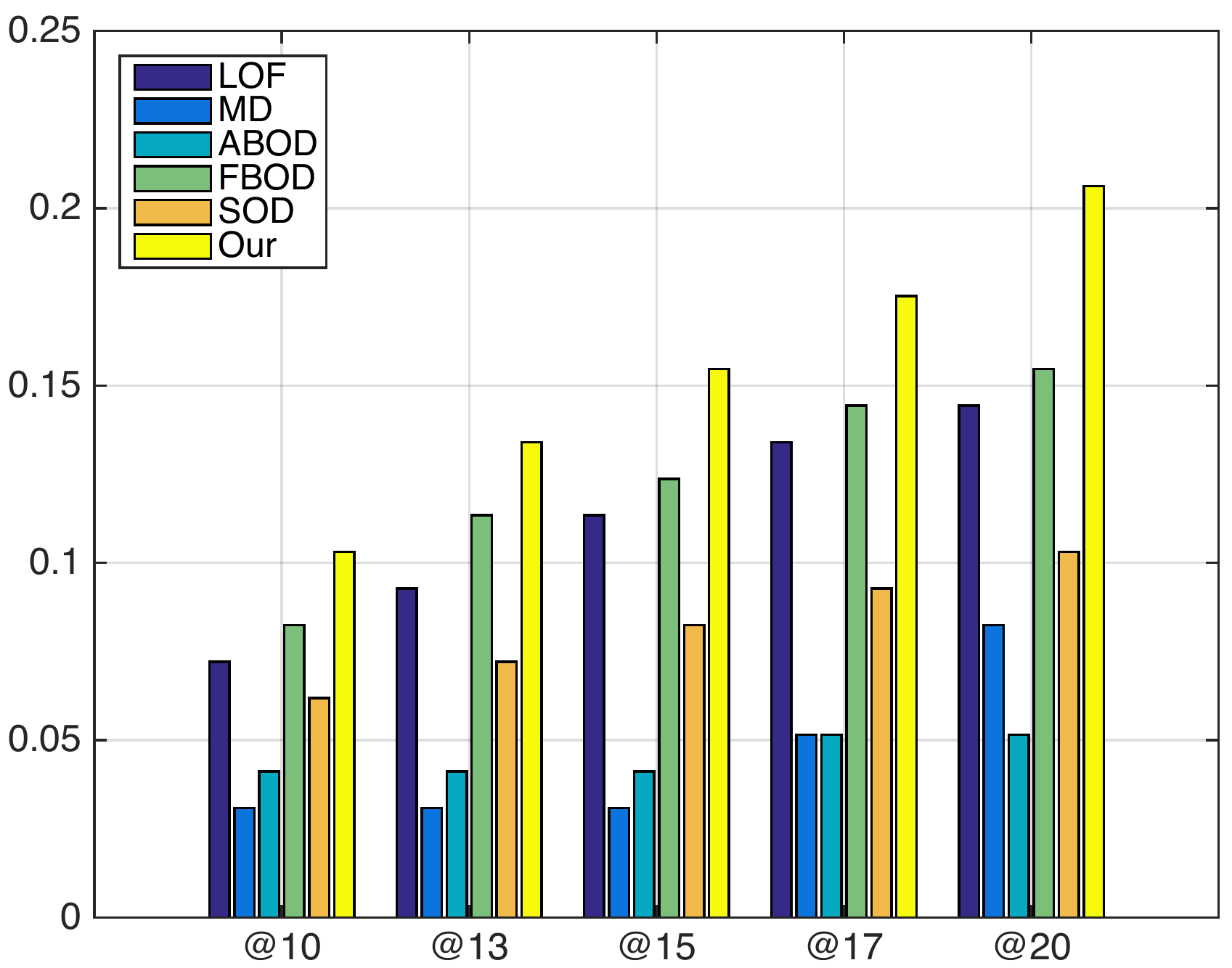}}
\subfigure[Fmeasure@N]{\includegraphics[width=43mm]{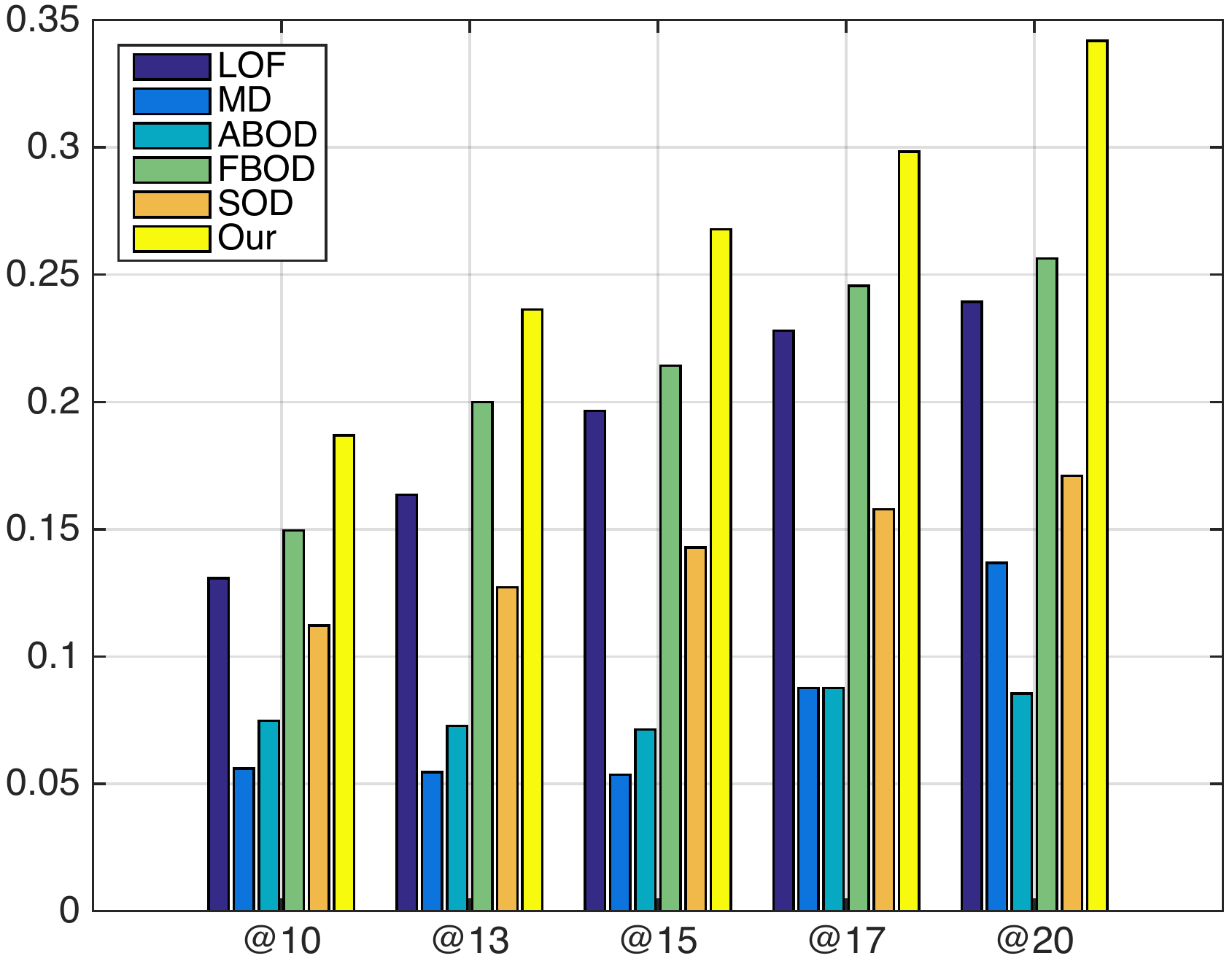}}
\vspace{-0.1 mm}
\caption{Comparison of REMIX with baseline algorithms on the sonar signals dataset}
\label{overall_performance_oc508}
\end{figure*}

\begin{figure*}[h]
\centering
\subfigure[Precision@N]{\label{ensemble_prec_oc541}\includegraphics[width=43mm]{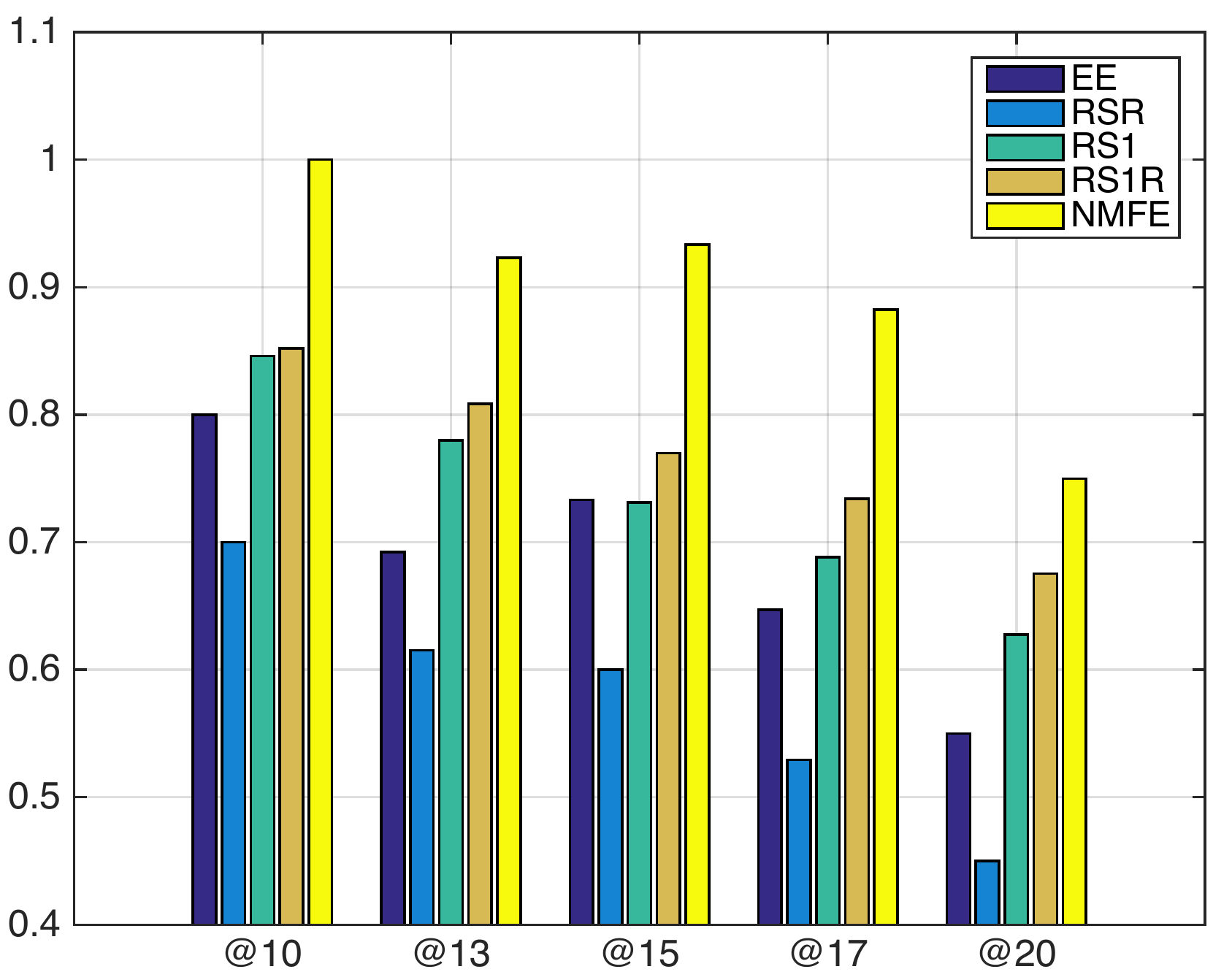}}
\subfigure[Recall@N]{\label{ensemble_recall_oc541}\includegraphics[width=43mm]{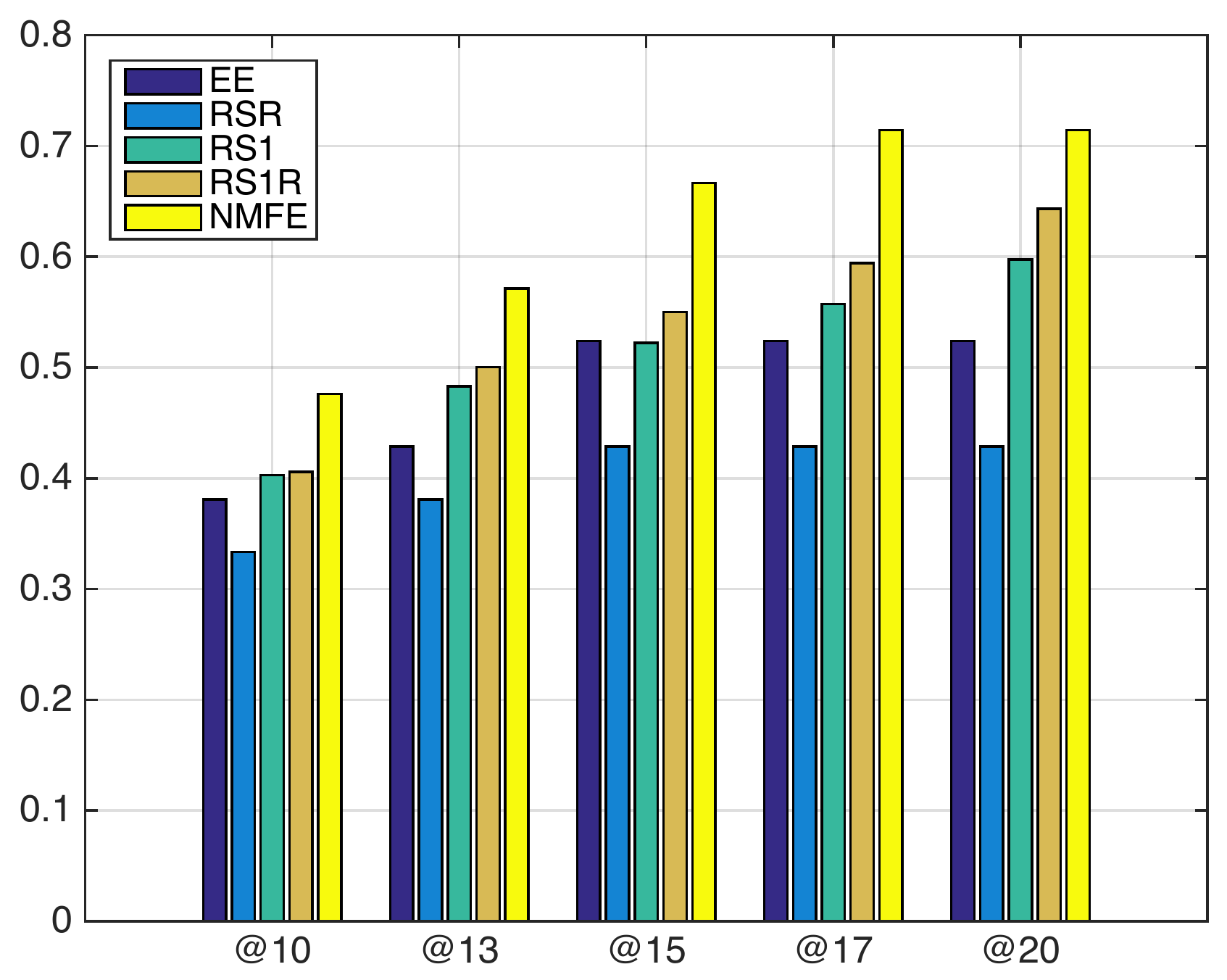}}
\subfigure[Fmeasure@N]{\label{ensemble_f_oc541}\includegraphics[width=43mm]{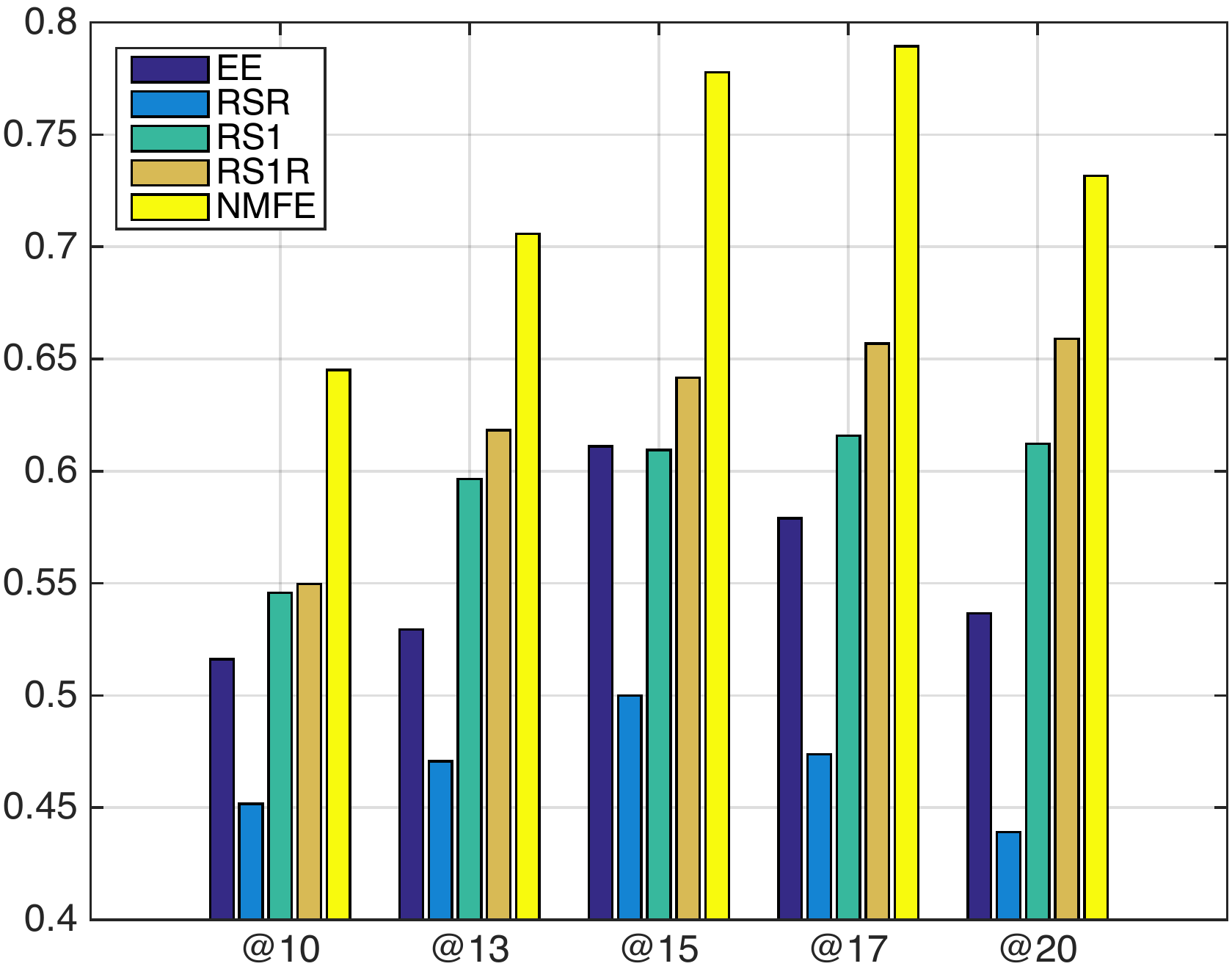}}
\vspace{-0.1 mm}
\caption{Comparison of REMIX with other exploration and ensemble strategies on the cardiac arrhythmia dataset}
\label{ensemble_performance_oc541}
\end{figure*}

\begin{figure*}[h]
\centering
\subfigure[Precision@N]{\label{ensemble_prec_oc508}\includegraphics[width=43mm]{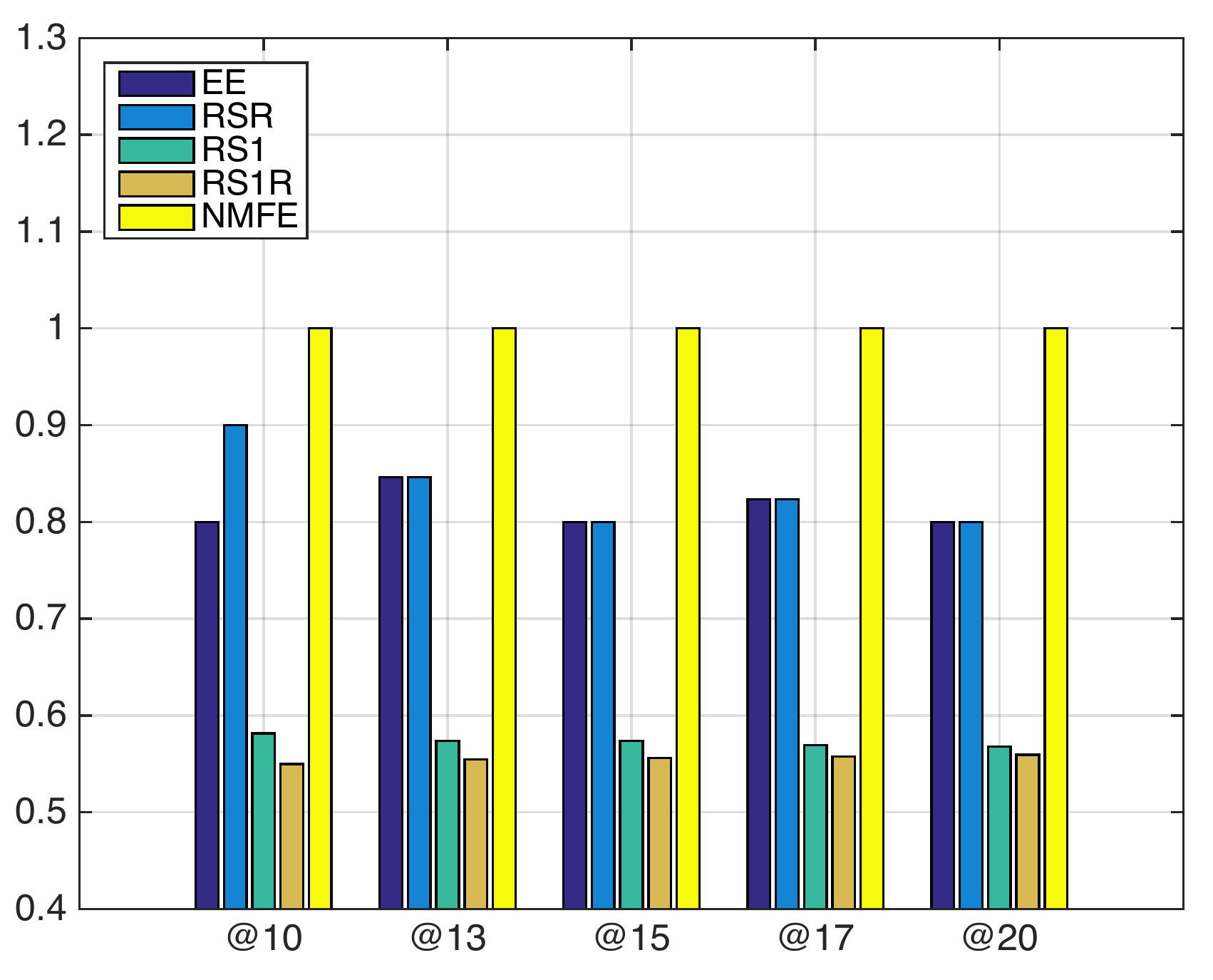}}
\subfigure[Recall@N]{\label{ensemble_recall_oc508}\includegraphics[width=43mm]{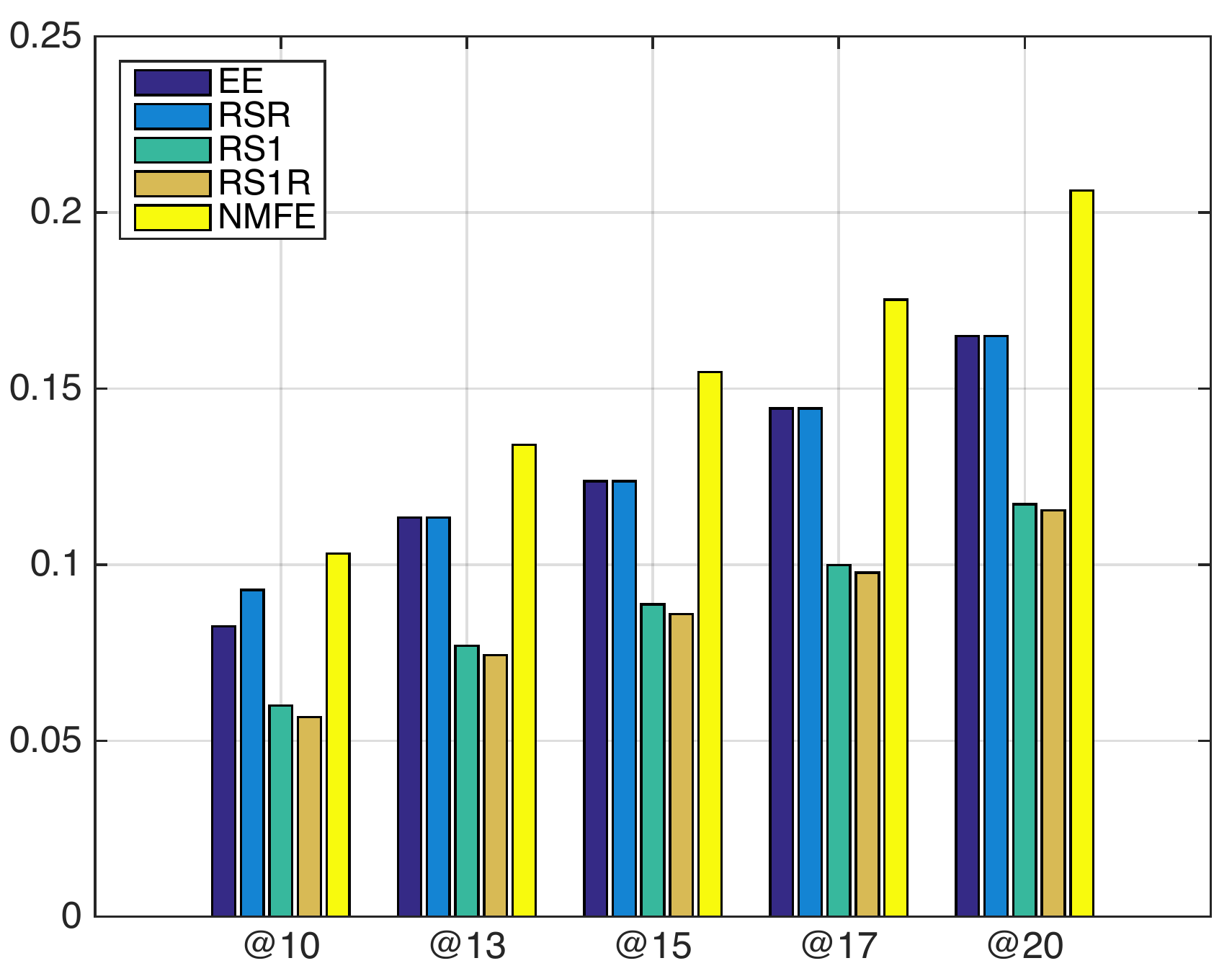}}
\subfigure[Fmeasure@N]{\label{ensemble_f_oc508}\includegraphics[width=43mm]{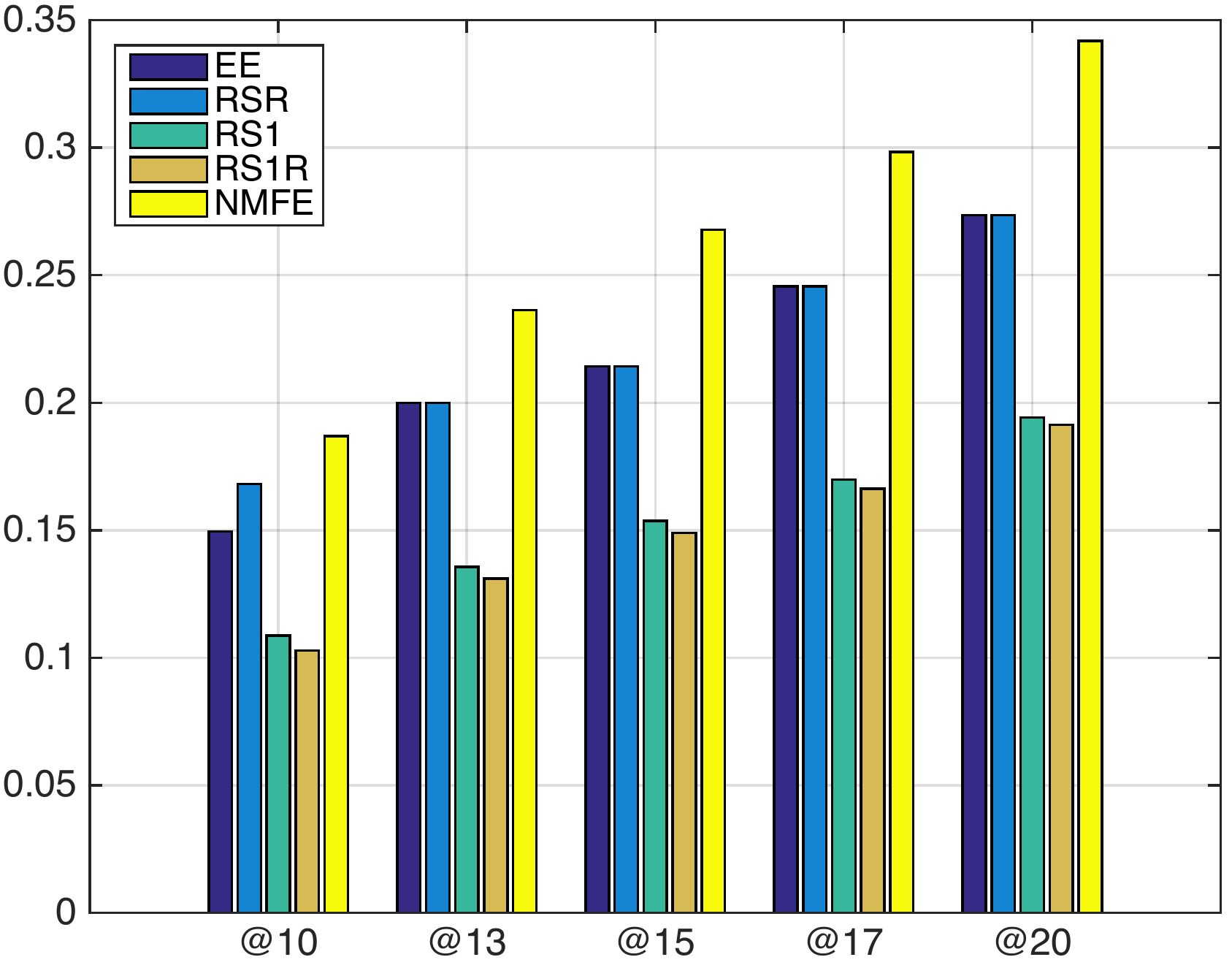}}
\vspace{-0.1 mm}
\caption{Comparison of REMIX with other exploration and ensemble strategies on the sonar signals dataset}
\label{ensemble_performance_oc508}
\end{figure*}

\begin{figure}[t]
\centering
\subfigure[Times costs on the cardiac arrhythmia dataset ]{\includegraphics[width=41mm]{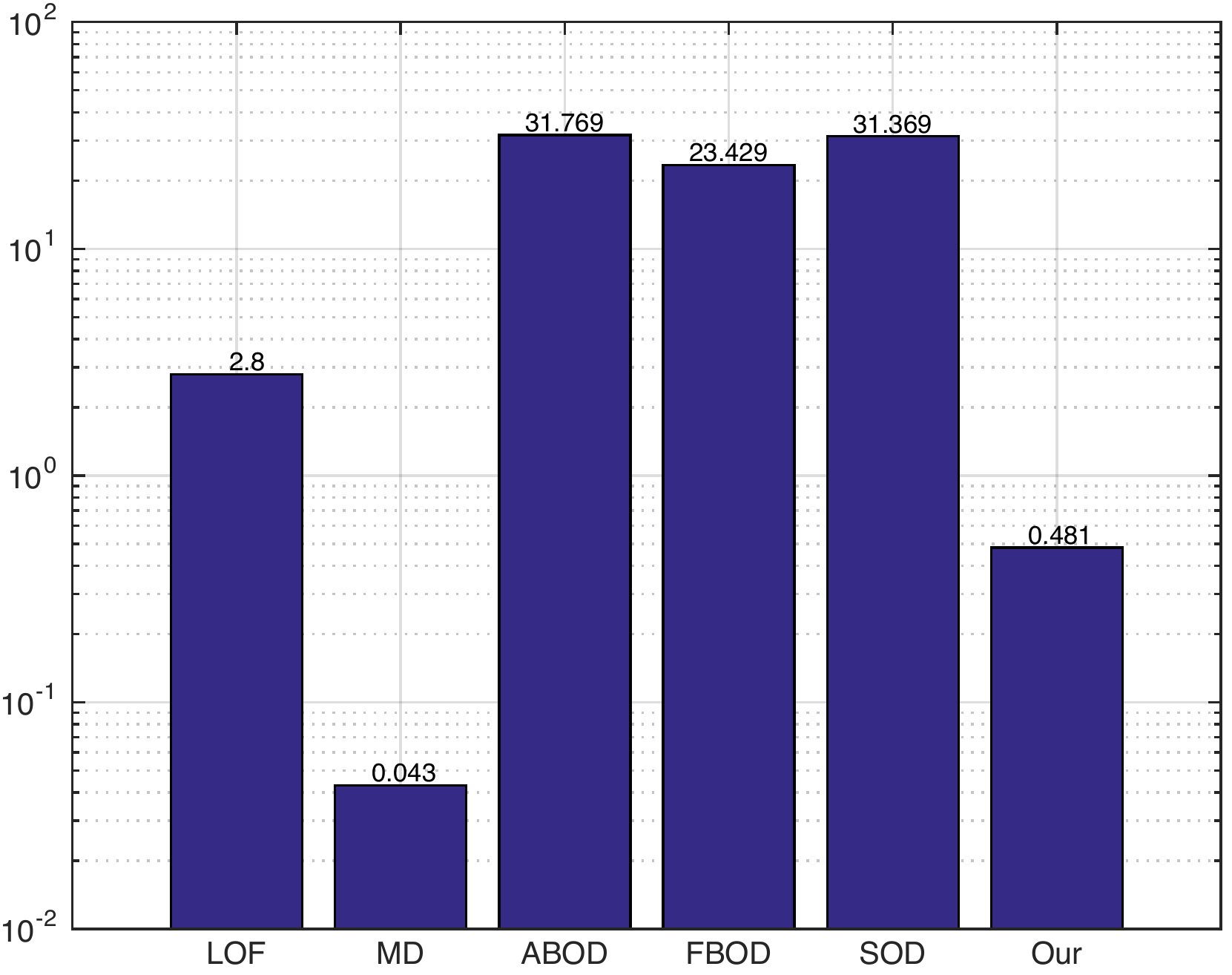}}
\subfigure[Times costs on the sonar signal dataset]{\includegraphics[width=41mm]{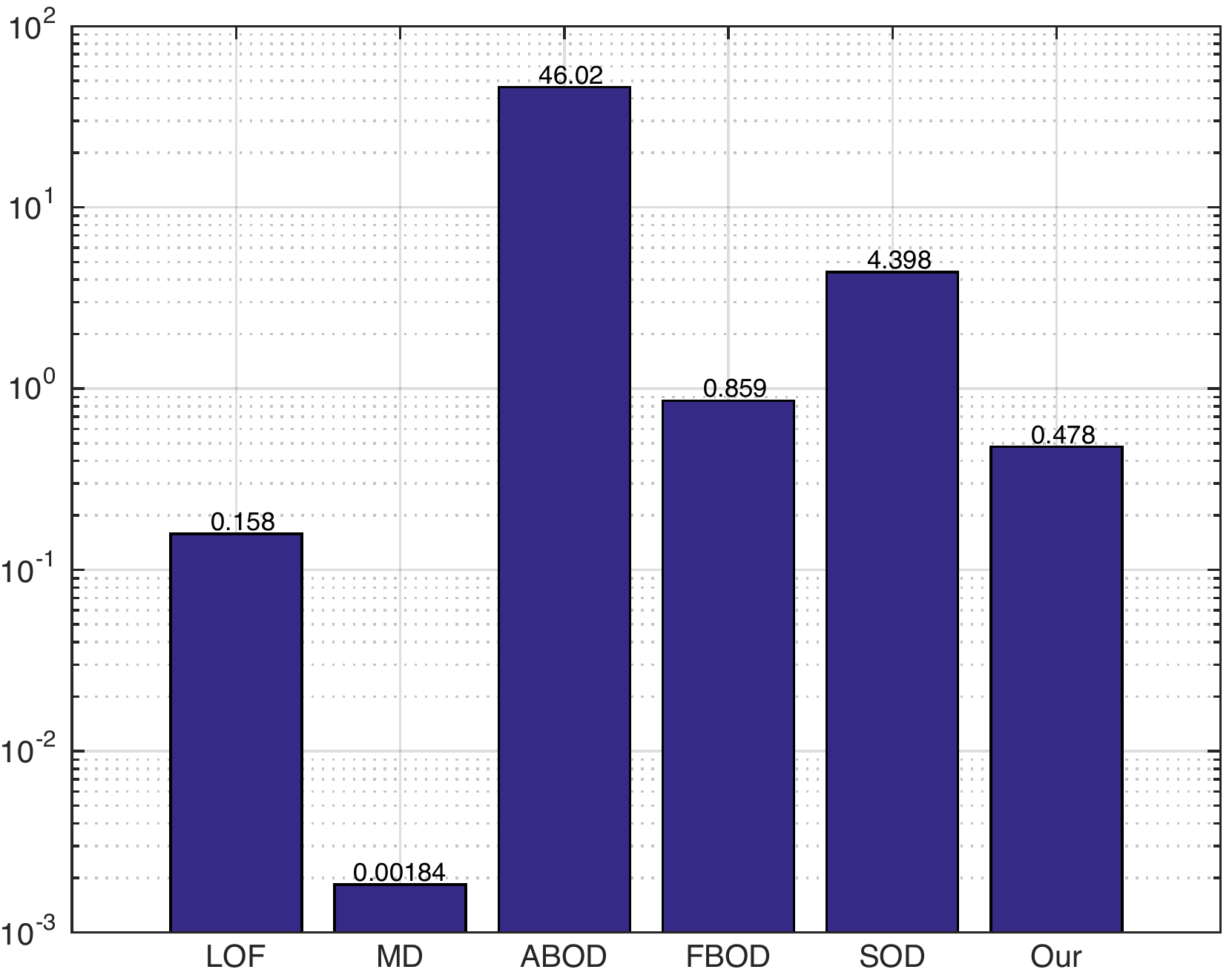}}
\caption{Comparison of execution costs of REMIX and baseline algorithms}
\label{my_times}
\end{figure}

\begin{figure}[ht]
\centering
\subfigure[Times costs on the cardiac arrhythmia dataset ]{\label{ensemble_times_541}\includegraphics[width=41mm]{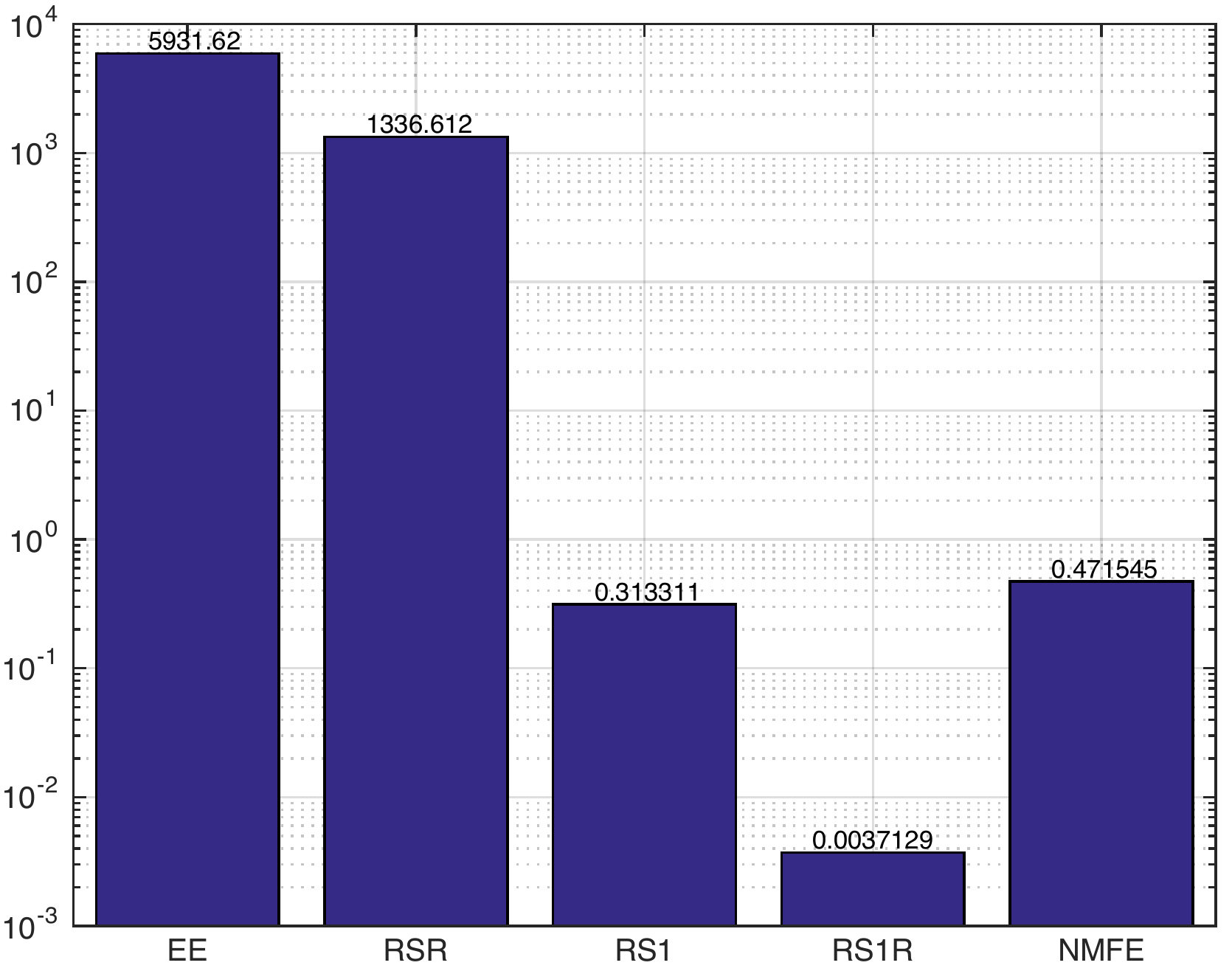}}
\subfigure[Times costs on the sonar signal dataset]{\label{ensemble_times_508}\includegraphics[width=41mm]{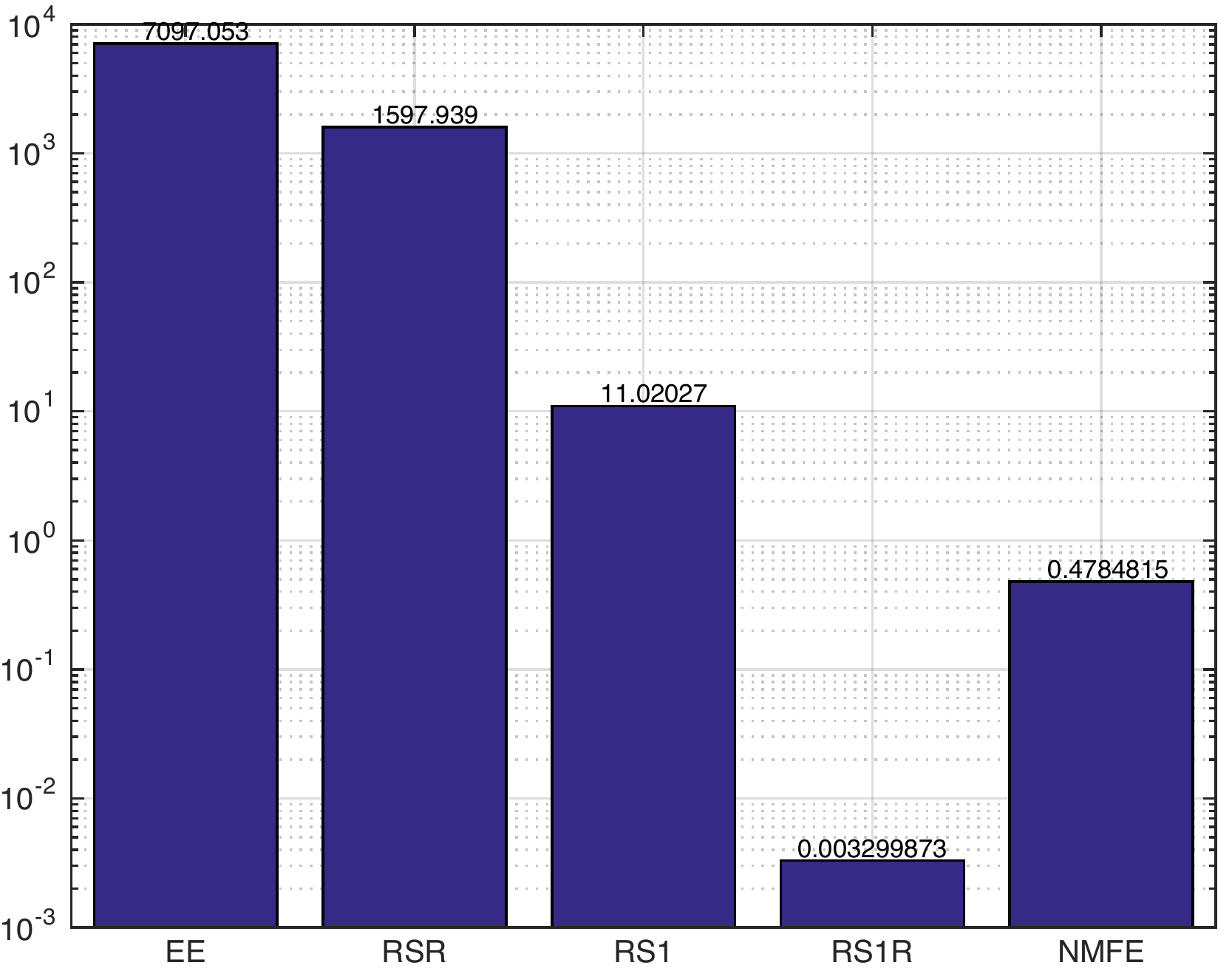}}
\caption{Comparison of execution costs of REMIX and other exploration strategies}
\label{ensemble_times}
\end{figure}

We now present a detailed study of the detection accuracy vs cost of REMIX on two datasets from the corpus.
The first dataset is cardiac arrhythmia. Irregularity in heart beat may be harmless or life threatening.
The dataset contains medical records like age, weight and patient's electrocardiograph related data of arrhythmia patients and outlier healthy people. The task is to spot outlier healthy people from arrhythmia patients.
The second dataset is about sonar signals bounced off of a metal cylinder.
The dataset contains outlier sonar signals bounced off a roughly cylindrical rock.
The task is to separate outlier rock related sonar signals from  cylinder-related sonar signals.

\subsubsection{Evaluation Metrics}
Recall that the REMIX perspective factorization scheme can be used as an outlier ensembling technique simply by setting the number
of perspectives to 1. We use REMIX in this ensembling mode for the rest of our experiments.
We now define Precision@N, Recall@N, and F-measure@N which we use to compare the detection accuracy of REMIX with various other approaches.
Let 1 denote outlier label and 0 denote the label for normal points in the expert labeled data.

{\bf Precision@N} Given the top-N list of data points $\mathcal{L}_{N}$ sorted in a
descending order of the predicted outlier scores, the precision is defined as:
$\mathrm{Precision}@N = \frac{|\mathcal{L}_{N} \bigcap \mathcal{L}_{\mathrm{= 1}}|}{N}$
where $\mathcal{L}_{\mathrm{= 1}}$ are the data points with expert outlier label = 1.

{\bf Recall@N} Given the top-N list of data points $\mathcal{L}_{N}$ sorted in a
descending order of the predicted outlier scores, the recall is defined as:
$\mathrm{Recall}@N = \frac{|\mathcal{L}_{N} \bigcap \mathcal{L}_{\mathrm{= 1}}|}{|\mathcal{L}_{\mathrm{= 1}}|}$
where $\mathcal{L}_{\mathrm{= 1}}$ are the outlier data points with label = 1.

{\bf F-measure@N} F-measure@N incorporates both precision and recall in a single metric by taking their harmonic mean:
$
\mathrm{F}@N= \frac{2\times \mathrm{Precision}@N \times \mathrm{Recall}@N}{\mathrm{Precision}@N + \mathrm{Recall}@N}
$

\subsubsection{Baseline Algorithms}
We report the performance comparison of REMIX vs baseline algorithms on the sonar signals dataset and the
cardiac arrhythmia dataset in terms of Precision, Recall, and F-measure.
In this experiment, we provide sufficient time for the baseline algorithms in $\{LOF, MD, ABOD, FBOD, SOD\}$
to complete their executions. Meanwhile, we set a limited exploration time budget of 0.5 second for REMIX.

{\bf Results on Effectiveness Comparison.}
Figure \ref{overall_performance_oc541} shows that on the cardiac arrhythmia dataset,
REMIX outperforms the five baseline algorithms in terms Precision@N,  Recall@N, and Fmeasure@N (N=10, 13, 15, 17,  20).
Figure \ref{overall_performance_oc508} shows that on the sonar signal dataset,
our method is consistently better than the baseline algorithms in terms Precision@N, Recall@N, and Fmeasure@N (N=10, 13, 15, 17,  20).

{\bf Results on Efficiency Comparison.}
Figure \ref{my_times} jointly show that on the cardiac arrhythmia dataset and the sonar signal dataset,
Mahalanobis distance (MD) takes the least time; Angle-based outlier detection (ABOD) takes the most time as angle is
expensive to compute; REMIX falls in the middle of this cost spectrum.

\subsubsection{The Effectiveness and Efficiency of REMIX NMFE}
We study the effectiveness of the REMIX NMFE ensembling strategy, in which we used the rank-1 NMF to
factorize the outlier score matrix $\Delta \approx \Lambda\Omega^{\top}$ and treat $\Omega$ as the predicted ensemble outlier scores.
Let $t$ denote the number of detectors selected in the REMIX MIP solution. Let $\mathcal{C}$ be the set of all candidate detectors enumerated by REMIX.
We compared REMIX with the following exploration and ensembling strategies:
\noindent{\bf (1) Exhaustive Ensemble (EE):} we execute all the detectors in $\mathcal{C}$ and averaged all the outlier scores;
\noindent{\bf (2) Randomly select $t$ detectors (RSR):} we randomly select and execute $t$ detectors from $\mathcal{C}$,
and then average the outlier scores of these detectors;
\noindent{\bf (3) Randomly select 1 detector (RS1):} we randomly select one detector and use its results;
\noindent{\bf (4) Randomly select 1 detector in the MIP solution (RS1R):} we randomly select one detector in the MIP solution and use its results;
In this experiment, we provide sufficient time for all strategies complete their execution.
We set a limited time budget of 0.5 seconds for REMIX.

{\bf Results on Effectiveness Comparison.}
Figure \ref{ensemble_performance_oc541} shows that on the cardiac arrhythmia dataset our strategy outperforms
the other exploration and ensembling strategies in terms of Precision@N,  Recall@N,
and Fmeasure@N (N=10, 13, 15, 17,  20). Figure \ref{ensemble_performance_oc508} shows that on the sonar signal dataset,
REMIX is consistently better than the other strategies in terms of Precision@N,
Recall@N, and Fmeasure@N (N=10, 13, 15, 17,  20).

{\bf Results on Efficiency Comparison.}
Figure \ref{ensemble_times_541} shows that on the cardiac arrhythmia dataset REMIX takes 0.48 second,
and EE, RSR, RS1 require much more time. While RS1R takes only 0.0037 second, its detection accuracy is lower than ours.
Figure \ref{ensemble_times_508} shows on the sonar signal dataset, our strategy takes only 0.49 second,
which is much less than the time costs of EE, RSR, RS1. While RS1R takes only 0.0033 second,
our method outperforms RS1R with respect to detection accuracy.

\section{Conclusions and Future Directions}
\label{sec:conclusion}
In this paper, we presented REMIX, a modular framework for outlier exploration which is the first to
study the problem of outlier detection in an interactive setting. At the heart of REMIX is an
optimization approach which systematically steers the selection of base outlier detectors in a manner
that is sensitive to their execution costs, while maximizing an aggregate utility function of the solution and also ensuring
diversity across algorithmic choice points present in exploration. Data analysts are naturally interested
in extracting and interpreting outliers through multiple detection mechanisms since
distinct outliers could lead to distinct actionable insights within the application domain.
REMIX facilitates this understanding in a practical manner by shifting the burden of exploration away from the analyst
through automation, and by summarizing the results of automated exploration into a few coherent heatmap visualizations called perspectives.

We believe many of the techniques presented in this paper could be of independent interest to other machine learning
problems. We are interested in extending the REMIX exploratory approach beyond outlier detection to clustering of high-dimensional data.
Another interesting direction of research is sequential recommendations for visual outlier exploration: in particular, we
are interested in approaches
for presenting outlier insights approximately through a sequence of visualizations (like heatmaps) such
that both the length of this sequence as well the perceptual error involved across the visualizations is minimized
while coverage across the various exploratory choice points is maximized. Also the study of outlier aspect mining in an interactive
setting -- which
deals with the inverse problem of finding explanatory features which characterize a given set of points as outliers in a
cost sensitive manner -- presents a new and interesting direction of research.

\bibliographystyle{plain}

\end{document}